\documentclass{article} 
\usepackage{iclr2027_conference,times}
\iclrfinalcopy  

\usepackage{amsmath,amsfonts,bm}

\def\eqref#1{equation~\ref{#1}}

\def\1{\bm{1}}

\DeclareMathAlphabet{\mathsfit}{\encodingdefault}{\sfdefault}{m}{sl}
\SetMathAlphabet{\mathsfit}{bold}{\encodingdefault}{\sfdefault}{bx}{n}

\usepackage[most]{tcolorbox}
\usepackage{multirow} 
\usepackage{booktabs}
\usepackage{hyperref}
\usepackage{url}
\usepackage{algorithm}
\usepackage{amsmath, amssymb, bm}
\usepackage{wrapfig}
\usepackage{algorithmic}
\usepackage{graphicx}
\usepackage[table]{xcolor}
\definecolor{clrVanilla}{RGB}{224,224,224}  
\definecolor{clrPink}{RGB}{250,224,226}     
\definecolor{clrYellow}{RGB}{252,247,206}   
\definecolor{clrBlue}{RGB}{219,226,244}     
\definecolor{clrOurs}{RGB}{223,240,224}
\definecolor{ablationpurple}{HTML}{EDE4F5}
\usepackage{pifont}
\usepackage{xcolor}
\usepackage{caption}

\title{RPSD: Retrieval-Privileged On-Policy Self-Distillation with Optimal Transport}

\author{
\begin{minipage}{0.95\textwidth}
\raggedright
\text{Dongyue Li}$^{1}$,
\text{Zebang Cheng}$^{2}$,
\text{Zhongyi Deng}$^{3}$,
\text{Hongbo Xu}$^{4}$,
\text{Yi Bin}$^{5}$,
\textbf{Fei Ma}$^{\dagger,4}$\\[4pt]
$^{1}$ Nagoya University \quad
$^{2}$ Shenzhen University\\
$^{3}$ Sichuan Agricultural University\\
$^{4}$ Guangdong Laboratory of Artificial Intelligence and Digital Economy\\
$^{5}$ Tongji University\\[4pt]
\end{minipage}
}

\begin{document}

\maketitle
\lhead{}
\renewcommand{\headrulewidth}{0pt}
\begingroup
\renewcommand{\thefootnote}{\fnsymbol{footnote}}
\footnotetext[2]{Corresponding author.}
\endgroup
\newcommand{\method}{\textsc{RPSD}}

\begin{abstract}
Sequential recommendation predicts a user's next interaction from historical behavior. Recent LLM-based methods recast this task as generative recommendation and increasingly adopt reinforcement learning such as GRPO for post-training. However, these methods suffer from two limitations, namely \textbf{\textit{credit-blind optimization}}, where a sequence-level reward is shared across all generated tokens, and \textbf{\textit{collaborative-blind supervision}}, where each training signal is derived from a single user and ignores behavioral patterns shared across similar users. To address both issues, we propose \textbf{R}etrieval-\textbf{P}rivileged on-policy \textbf{S}elf-\textbf{D}istillation (RPSD). For credit-blind optimization, we introduce on-policy self-distillation (OPSD), in which a teacher with privileged knowledge re-evaluates the student's on-policy response, and the token-level confidence difference between the student and teacher reshapes the shared sequence-level advantage into token-specific credit. However, the privileged knowledge in vanilla OPSD is still derived from a single user, leaving collaborative-blind supervision unresolved. We therefore ground the teacher in collaborative behavioral evidence from similar users. Specifically, we use temporally aware optimal transport (OT) to compare interaction trajectories across users by jointly considering item similarity and temporal structure, and further apply diversity-aware selection to retain complementary trajectories as privileged evidence. In this way, on-policy self-distillation provides fine-grained token-level credit, while OT enriches the teacher with structured cross-user evidence. To the best of our knowledge, this is the first work to bring on-policy self-distillation to sequential recommendation and to use cross-user behavior as its privileged signal. Extensive experiments across multiple real-world datasets demonstrate that RPSD achieves state-of-the-art performance.
\end{abstract}

\section{Introduction}

Sequential recommendation predicts the next item that a user will interact with from
the ordered history of past interactions and it is central to content,
e-commerce, and media platforms \cite{hug-vis} where user interests change over
time~\citep{gru4rec,sasrec}. Traditional methods treat the sequential recommendation problem as next-item
prediction over an interaction sequence, evolving from recurrent and
self-attention encoders~\citep{gru4rec,sasrec} to graph-based and contrastive
variants, which model higher-order transitions and augment sparse
signals~\citep{lightgcn,cl4srec,srepdegen}. Although these models are effective and
efficient, they represent items as opaque identifier embeddings and reason
over co-occurrence statistics rather than the semantics of what a user consumed,
which limits generalization when histories are short. An ideal recommender,
however, should do more than name which candidate comes next and it should ground
that choice in why the recommended item matches the user's evolving interest.

Large language models (LLMs) address the semantic side of this gap by casting
recommendation as language generation, either unifying recommendation tasks in a
text-to-text format~\citep{p5,tiger,gnprsid,lsig,minionerec} or fine-tuning a pretrained LLM to reason over
item titles and name the next item directly in
text~\citep{tallrec,llara,collm}. To align such an LLM with the recommendation objective, reinforcement learning
has become a standard post-training recipe. Recent work adapts GRPO~\citep{deepseekmath} to LLM-based recommendation, tailoring its reward and credit assignment to the ranking structure of the task. For example,
Rank-GRPO~\citep{rankgrpo} moves the reward from the whole sequence to each
individual rank, so that later positions in the list get their own credit instead
of sharing one signal. RecZero~\citep{reczero} keeps GRPO unchanged but reshapes the
rollout into explicit analyze-user, analyze-item, match, and rate steps, replacing
exact-match correctness with a reward graded by the predicted rating's closeness to
the true one. However, these adaptations only change which part of a response the reward is applied to.
Rank-GRPO gives each rank its own reward and RecZero gives each reasoning step its own, but in
both cases a single number is still shared by all the tokens inside that rank or step. The source of the reward is left
untouched as well. Whether it is computed over the sequence, the rank, or the step, it always
comes from the next item of the user being scored. A reward of this kind can confirm that a
recommendation is right, but one user's interactions are a thin basis for learning why, since a
short history reflects accident as much as preference. The same pattern observed across many
users with comparable histories is far more dependable, and it is the only evidence available
when a single history is too short to be read on its own. Therefore, two problems remain unaddressed, namely \textbf{\textit{credit-blind optimization}}
and \textbf{\textit{collaborative-blind supervision}}.

\begin{figure*}
\includegraphics[width=\textwidth]{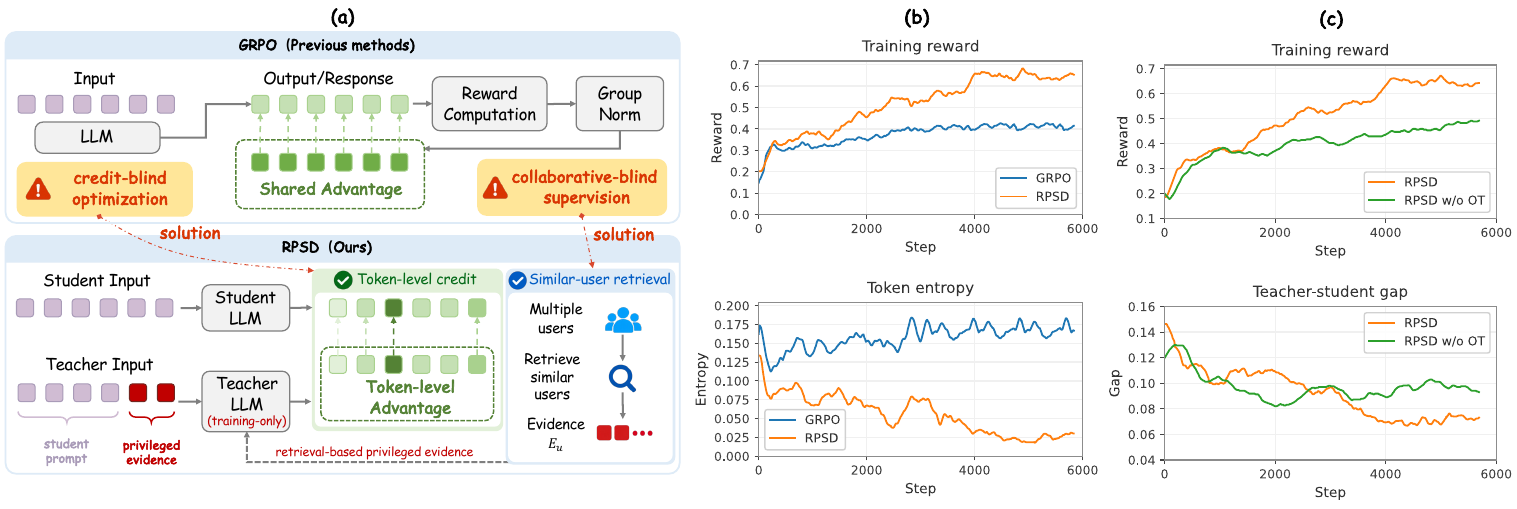}
  \vspace{-0.3in}\caption{\textbf{RPSD combines token-level credit assignment with retrieved evidence.}
(a) Conceptual differences between RPSD and GRPO.
(b) Training reward and token entropy of GRPO and RPSD.
(c) Comparison between RPSD and RPSD w/o OT, which only takes the ground-truth item as privileged knowledge of the teacher.}
\label{motivation}
\end{figure*}

\textbf{\textit{Credit-blind optimization}} concerns how the reward is assigned. Not every token in a response contributes equally to the recommendation. Most tokens provide shared linguistic or formatting structure, while only a small subset determines which item is ultimately selected. GRPO nevertheless assigns the same sequence-level advantage to every generated token, providing no indication of which tokens actually make the recommendation correct. On-policy self-distillation (OPSD)~\citep{opsd} provides a promising direction for addressing this limitation. By deriving token-level guidance from a more informed teacher on the same on-policy response, OPSD enables finer-grained credit assignment while preserving the original policy and inference procedure. These properties make it particularly appealing for sequential recommendation. However, the effectiveness of OPSD is ultimately limited by the informativeness of the teacher's privileged knowledge. If the teacher lacks sufficient behavioral context, the resulting token-level guidance may still be inadequate for capturing the preference signals underlying the target item. Vanilla OPSD typically relies on the ground-truth label as the privileged signal, which tells the model \emph{which} candidate is correct but provides limited information about \emph{why} that candidate matches the user's preference.
Therefore, the second issue \textbf{\textit{collaborative-blind supervision}} is introduced. Existing generative recommenders derive supervision from the user alone, overlooking regularities shared across similar users.

To address these two limitations, we propose \textbf{R}etrieval-\textbf{P}rivileged on-policy \textbf{S}elf-\textbf{D}istillation (\method{}), which combines OPSD with temporally aware OT. \method{} first constructs cross-user behavioral evidence from the training corpus. Specifically, we use temporally aware OT to compare interaction trajectories across users by jointly modeling item similarity and temporal structure, followed by diversity-aware selection to retain complementary trajectories. Their subsequent interactions, together with the ground-truth item, form the privileged knowledge available to the teacher.
The teacher then re-evaluates the student's on-policy response under this privileged view, and the resulting token-level confidence difference is used to reshape the sequence-level advantage into token-specific credit. The privileged knowledge is used only during training, leaving inference unchanged and introducing no additional deployment cost. As shown in Figure~\ref{motivation}, the retrieved behavioral evidence further strengthens teacher supervision and leads to improved training dynamics. To the best of our knowledge, \method{} is the first method to introduce OPSD into sequential recommendation and to use cross-user behavior as privileged knowledge. Our contributions are summarized as follows:

\textbf{(1)} RPSD, a novel on-policy self-distillation framework, is proposed for sequential recommendation to enable fine-grained token-level credit assignment through privileged teacher guidance.

\textbf{(2)} A temporally aware optimal transport mechanism is introduced to align similar user trajectories, from which cross-user evidence is constructed as privileged knowledge.

\textbf{(3)} Extensive experiments on three benchmark datasets are conducted to demonstrate the effectiveness of RPSD and its consistent superiority over strong state-of-the-art baselines.

\section{Related Work}

\subsection{Reinforcement Learning for LLM-based Recommendation}
Casting recommendation \citep{gnprsid,lsig,gr-handbook,minionerec,emb-sid} as language generation lets LLMs reason over item semantics
rather than opaque identifiers~\citep{p5,tiger,tallrec,llara,collm,llm4recsurvey},
and reinforcement learning with verifiable rewards has become a standard recipe for
aligning such models with the recommendation
objective~\citep{deepseekmath,recr1,reczero,recllmenv,easyrl4rec}. Since final-answer rewards are sparse,
lots of works adapt GRPO to the ranking structure of the task. Rank-GRPO
makes each rank, rather than the token or the whole sequence, the unit of reward to
remove non-causal credit assignment~\citep{rankgrpo}, and RecZero restructures the
rollout into explicit reasoning steps supervised by a graded rule-based
reward~\citep{reczero}. Parallel works pursue the same goal through preference
objectives, where S-DPO extends DPO with multiple negatives~\citep{sdpo} and C-APO
augments the revealed-preference ordering with a coherence-based ordering to
down-weight noisy user choices~\citep{capo}. These methods differ in what unit the
reward is computed over and how the preference ordering is shaped, yet the signal
they supply still scores whether a recommendation is right rather than which tokens
carry the evidence for it. We instead densify the signal to the token level while grounding the label in retrieved behavior.

\subsection{On-policy Self-Distillation}
On-policy distillation~\citep{opd} (OPD) supervises a student on its own sampled trajectories with
dense token-level targets from a teacher, combining the density of distillation with
the distributional match of on-policy learning~\citep{minillm}. OPSD removes the external teacher by letting a single model act as both
student and teacher under different contexts, where the teacher conditions on
privileged information the student cannot see and re-scores the student's tokens,
turning a sparse outcome reward into dense supervision~\citep{opsd}. A recognized limitation, however, is that conditioning on an
instance-level label lets the teacher rationalize the answer without grounding it in
the input, collapsing its signal into a shortcut~\citep{omniopsd}. These extensions
all draw their privilege from a label or a stronger model, and none targets
recommendation. To the best of our knowledge, we are the first to introduce OPSD to sequential recommendation and to ground label-based privileged information in retrieved behavioral evidence from similar users.

\subsection{Retrieval-Augmented Recommendation and Knowledge Distillation}
Retrieval-augmented recommendation enriches
sparse histories with behaviorally similar users, where ReLLa~\citep{rella} retrieves relevant
behaviors to improve long-history comprehension and RALLRec~\citep{rallrec}
strengthens the retrieval through representation learning. However, the retrieved neighbors are fed to these two recommenders at inference, so retrieval
quality directly shapes the deployed prediction. We instead use the continuations of
similar users only as a training-time privileged signal, leaving inference to the
ordinary prompt. Knowledge distillation transfers a teacher's ranking ability to a smaller student, as in RCE-KD~\citep{rcekd},
which reconnects the cross-entropy distillation loss for tighter ranking
transfer. Our RPSD uses the same model as both teacher and student, with the teacher additionally conditioned on retrieved behavioral evidence and the ground-truth item to provide privileged supervision only during training.

\section{Method}
\label{sec:method}

We present \method{}, a retrieval-privileged on-policy self-distillation
framework for LLM-based sequential recommendation. The student samples a
recommendation from the task prompt, while a teacher, which is the same model
under a privileged context, re-scores the same tokens and provides dense
token-level credit. The privileged context combines two pieces of training-only information: the ground-truth next item, which tells the teacher \emph{which} candidate is correct, and behavioral evidence retrieved from similar users, which grounds \emph{why} it fits the current history.
 \begin{figure*}
  \includegraphics[width=\textwidth]{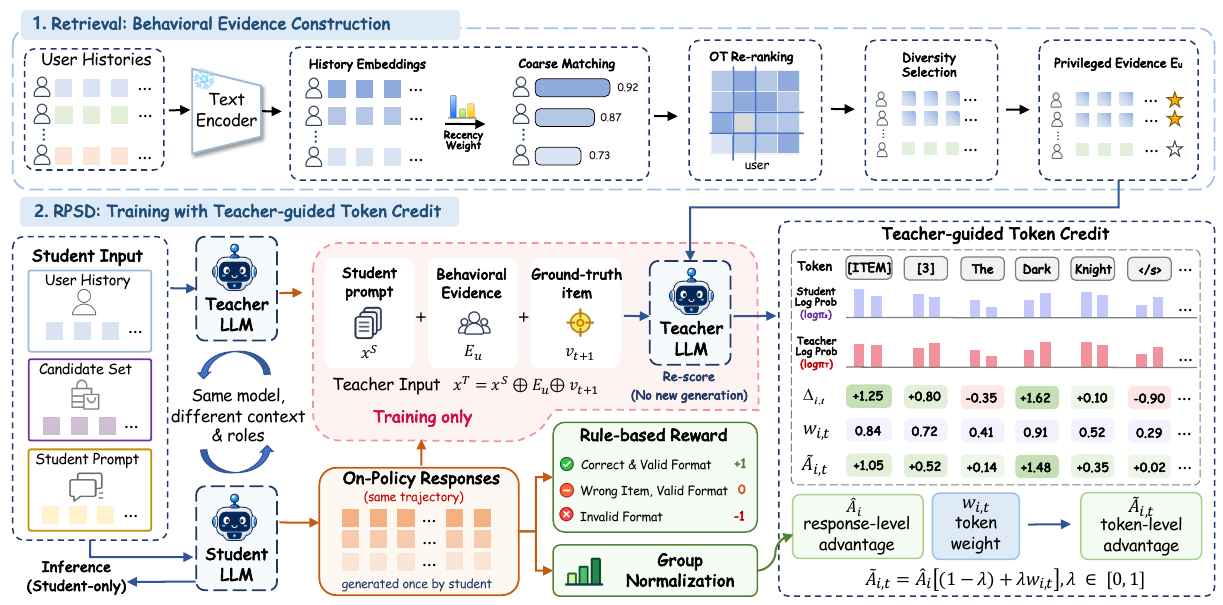}
  \vspace{-0.3in} \caption{\textbf{Overview of RPSD.} The student samples on-policy responses from the original recommendation prompt, while a retrieval-privileged teacher re-scores the same trajectory using behavioral evidence only during training to derive token-level credit.}
  \label{overview}
\end{figure*}
\subsection{Problem Formulation}
\label{sec:formulation}
 
In a general sequential recommendation scenario, we consider a user set
$\mathcal{U}$ and an item set $\mathcal{V}$. Given the interaction sequence
$S_u=\{v^u_1, v^u_2, \dots, v^u_t\}$ of a user $u\in\mathcal{U}$, where
$v^u_j\in\mathcal{V}$ is the item interacted with at step $j$, the goal is to
predict the next item $v^u_{t+1}$ the user is most likely to engage with.
 
We take the most recent $L$ interactions of
$S_u$ as the context $H_u$ and construct a candidate set
$\mathcal{C}_u=\{v^u_{t+1}\}\cup\mathcal{N}_u$ of size $n_c$, where
$\mathcal{N}_u$ contains negatives sampled from the user's non-interacted
items. We serialize $(H_u,\mathcal{C}_u)$ into a student prompt
$
x_u^{S} \;=\; \mathrm{Prompt}_S(H_u, \mathcal{C}_u)$. For
training, we additionally build a teacher prompt over the same instance that
exposes $x_u^{S}$ to retrieved behavioral evidence $E_u$ and the ground-truth item $v^u_{t+1}$, forming a privileged view used solely to shape the
training signal,
$
x_u^{T} \;=\; \mathrm{Prompt}_T(H_u, \mathcal{C}_u, E_u, v^u_{t+1})$.
Given $x_u^{S}$, the policy $\pi_\theta$ generates a response
$y=(y_1,\dots,y_T)$ that must follow the correct item format and predict the right next item
$v^u_{t+1}$. Training follows a two-stage paradigm, where the pretrained LLM is first adapted to the instruction format by SFT, after which \method{} is leveraged for policy optimization. At inference time, the model observes only $x_u^{S}$ and has no
access to $E_u$ or $v^u_{t+1}$.

\subsection{Retrieval-Based Privileged Evidence}
\label{sec:ot}

In addition to
revealing the label, we equip the teacher with behavioral evidence. For each
user, we retrieve similar interaction histories from \emph{other} users and
summarize their subsequent behavior, which the teacher reads as privileged
information.

Given the current user, we retrieve behaviorally similar history windows from a
pool built over all training users and read how those users continued.
Concretely, we slide a length-$K$ window over every training user sequence to
build a shared pool of history windows
$\mathcal{W}=\{(h_j,v_j^{+},u_j)\}_{j=1}^{M}$ with
$h_j=(v'_{j,1},\ldots,v'_{j,K})$, where $h_j$ is a user's recent history,
$v_j^{+}$ is the item that user watched next, and $u_j$ is the user. Each item is encoded from the \emph{title
text} into a unit vector $e_v\in\mathbb{R}^{d}$, and a length-$K$ history
$H=(v_0,\ldots,v_{K-1})$ is summarized by a recency-weighted vector
\begin{equation}
\bar e(H)=
\frac{\sum_{p} a_p \cdot e_{v_p}}
{\left\|\sum_{p} a_p \cdot e_{v_p}\right\|_2},
\qquad
a_p=
\frac{\exp\!\left[-\beta \cdot (K-1-p)\right]}
{\sum_{r}\exp\!\left[-\beta \cdot (K-1-r)\right]},
\label{eq:recency}
\end{equation}
where a larger $\beta$ assigns smaller weights to earlier interactions in the user's history, i.e., interactions that are farther from the most recent position, and $p=0$ denotes the earliest position. We use this vector only for a coarse recall stage that
selects top-$N_0$ windows by cosine
$\langle\bar e(H_u),\bar e(h_j)\rangle$, and then refine these shortlisted windows
by OT re-ranking.

Let $H_u^{K}=(v^u_{t-K+1},\dots,v^u_t)$ denote the $K$ most recent items of
$H_u$. We score each window $h_j$ by aligning $h_j$ with $H_u^{K}$
through unbalanced entropic OT. We use the
query mass $a=(a_0,\ldots,a_{K-1})$ from Eq. (\ref{eq:recency}), and the window
mass $b$ is computed by the same rule. The pairwise cost is defined as
\begin{equation}
C_{pq}=1-\langle e_{v_p},e_{v'_q}\rangle
+\gamma\cdot\left|\tfrac{p}{K-1}-\tfrac{q}{K-1}\right|,
\label{eq:cost}
\end{equation}
whose first term measures semantic dissimilarity and second term penalizes
temporally inconsistent alignments with $\gamma\ge0$. We then
solve
\begin{equation}
\mathrm{OT}_{\varepsilon,\rho}(a,b;C)=\min_{P\ge 0}\,\langle P,C\rangle
+\varepsilon\,\mathrm{KL}(P\,\|\,\mathbf{1}\mathbf{1}^{\top})
+\rho\big[\mathrm{KL}(P\mathbf{1}\,\|\,a)+\mathrm{KL}(P^{\top}\mathbf{1}\,\|\,b)\big]
\label{eq:uot}
\end{equation}
with the Sinkhorn algorithm, where $\varepsilon$ controls entropic smoothing and
$\rho$ relaxes the two marginals. To make the transport costs comparable across windows with different transported masses, we normalize the optimal transport plan $P^\ast$ by the transported mass and then map $P^\ast$ to
a similarity
$s(H_u^{K},h_j)=\exp\!\big(-\langle P^\ast,C\rangle/(\|P^\ast\|_1\,\tau)\big)$.
Finally, we greedily select a diverse set $\mathcal{S}$ of windows until
$|\mathcal{S}|=3$,
\begin{equation}
j^\ast=\arg\max_{j\notin\mathcal{S}}
\big[s_j-\lambda_{\mathrm{div}}\cdot\max_{k\in\mathcal{S}}
\langle\bar e(h_j),\bar e(h_k)\rangle\big],
\label{eq:mmr}
\end{equation}
where $\lambda_{\mathrm{div}}$ trades off relevance and diversity. Each selected window is turned into one line of natural language that lists the user's $K$ recent items, then the item they interacted with next, tagged with
the similarity $s_j$. Stacking these lines forms the evidence block $E_u$, which tells the
teacher how users with histories like the current one actually continued.

\subsection{Retrieval-Privileged on-policy Self-Distillation}
\label{sec:opsd}

Building on the retrieved evidence $E_u$ in Section~\ref{sec:ot}, we introduce
RPSD, which addresses the uniform credit of GRPO and the label-only privilege
of vanilla OPSD. Specifically, the \emph{same} model acts as both
student and teacher. The student samples responses from the ordinary prompt
$x^{S}$. The teacher re-scores the same responses under the privileged prompt
$x^{T}=x^{S}\oplus E_u\oplus v_{t+1}^{u}$, where the label tells it
\emph{which} candidate is correct and $E_u$ shows \emph{why} it fits the
current history. The resulting token-level confidence gap is used only to
redistribute the GRPO advantage across tokens, so the direction of each update
is still determined by the verifiable reward. Following
GRPO, we sample $G$ responses $\{o_1,\ldots,o_G\}$ under
$x^{S}$ and assign each a rule-based reward $r_i$: $+1$ for a valid format with
the correct item, $0$ for a valid format with a wrong item, and $-1$ for an
invalid format. The response-level advantage is the group-normalized reward
\begin{equation}
\widehat A_i=\frac{r_i-\operatorname{mean}(\{r_j\}_{j=1}^{G})}
{\operatorname{std}(\{r_j\}_{j=1}^{G})},
\label{eq:group_advantage}
\end{equation}
which standard GRPO shares uniformly across all tokens of $o_i$. We score every sampled token under both contexts with stop-gradient and take
their difference
\begin{equation}
\Delta_{i,t}=\log \pi_{\operatorname{sg}(\theta)}\!\left(o_{i,t}\mid x^{T},o_{i,<t}\right)
-\log \pi_{\operatorname{sg}(\theta)}\!\left(o_{i,t}\mid x^{S},o_{i,<t}\right),
\label{eq:dual_scoring}
\end{equation}
so a positive $\Delta_{i,t}$ means the evidence makes token $o_{i,t}$ more
plausible. Since the evidence shifts the whole sequence likelihood,
$\Delta_{i,t}$ is centered and clipped as
$\delta_{i,t}=\operatorname{clip}(\Delta_{i,t}-\tfrac{1}{T_i}\sum_{t'}\Delta_{i,t'},-c_\delta,c_\delta)$,
with stability threshold $c_\delta$.
We then convert this centered gap into an advantage-aware token weight
\begin{equation}
w_{i,t}=\operatorname{clip}\!\left(\exp\!\left[\operatorname{sign}(\widehat A_i)\cdot\delta_{i,t}\right],
\,1-\epsilon_w,\,1+\epsilon_w\right),
\label{eq:token_weight}
\end{equation}
so that tokens with stronger teacher support are reinforced more when
$\widehat A_i>0$, and tokens the teacher also disfavors receive stronger
negative credit when $\widehat A_i<0$, with $\epsilon_w$ bounding the guidance
strength. The
shared advantage is then refined into a token-specific one
$
\widetilde A_{i,t}=\widehat A_i\left[(1-\lambda)+\lambda w_{i,t}\right]$,
where $\lambda\in[0,1]$ sets the strength of privileged guidance, and
$\lambda=0$ or $\epsilon_w=0$ recovers standard GRPO. The final objective of our RPSD is
\begin{equation}
\mathcal{L}_{\mathrm{\method{}}}
=-\,\mathbb{E}\!\left[\frac{1}{G}\sum_{i=1}^{G}\frac{1}{T_i}\sum_{t=1}^{T_i}
\min\!\left(\rho_{i,t}\cdot\widetilde A_{i,t},\,
\operatorname{clip}(\rho_{i,t},1-\epsilon,1+\epsilon)\cdot\widetilde A_{i,t}\right)\right]
+\beta_{\mathrm{KL}}\cdot\mathcal{L}_{\mathrm{KL}},
\label{eq:opsd_objective}
\end{equation}
where $\rho_{i,t}=\pi_{\theta}(o_{i,t}\mid x^{S},o_{i,<t})/
\pi_{\theta_{\mathrm{old}}}(o_{i,t}\mid x^{S},o_{i,<t})$ is the token-level
importance ratio, $\epsilon$ is the clipping range, and
$\mathcal{L}_{\mathrm{KL}}$ regularizes toward the reference model as in GRPO. At inference,
the model conditions solely on $x^{S}$, so the evidence is used only during
training to reshape token credit.

\begin{table}[t]
\centering
\caption{Recommendation performance comparisons on three different datasets ($n_c=10$). Bold indicates the
best results among all methods.}
\label{tab:main_results}
\renewcommand{\arraystretch}{1.15}
\resizebox{\linewidth}{!}{
\begin{tabular}{cl cccc cccc cccc}
\toprule
& \multirow{2}{*}{\textbf{Model}} &
\multicolumn{4}{c}{\textbf{MovieLens-1M}} &
\multicolumn{4}{c}{\textbf{MicroLens}} &
\multicolumn{4}{c}{\textbf{Netflix}} \\
\cmidrule(lr){3-6}\cmidrule(lr){7-10}\cmidrule(lr){11-14}
& & HR@3 & HR@5 & NG@3 & NG@5 & HR@3 & HR@5 & NG@3 & NG@5 & HR@3 & HR@5 & NG@3 & NG@5 \\
\midrule
\multirow{8}{*}{\textbf{Non-LLM}}
& \textbf{GRU4Rec} \citep{gru4rec}  & 0.5507 & 0.7131 & 0.4762 & 0.5236 & 0.5080 & 0.6440 & 0.4369 & 0.4756 & 0.5152 & 0.7152 & 0.3659 & 0.4494 \\
& \textbf{SASRec} \citep{sasrec} & 0.6073 & 0.7498 & 0.5090 & 0.5638 & 0.5243 & 0.6597 & 0.4439 & 0.4958 & 0.5560 & 0.7570 & 0.4042 & 0.4786 \\
& \textbf{LightGCN} \citep{lightgcn} & 0.5808 & 0.7167 & 0.4894 & 0.5330 & 0.5138 & 0.6604 & 0.4344 & 0.4913 & 0.5430 & 0.7420 & 0.3928 & 0.4702 \\
& \textbf{CL4SRec} \citep{cl4srec}  & 0.6221 & 0.7557 & 0.5306 & 0.5861 & 0.5482 & 0.6826 & 0.4694 & 0.5186 & 0.5788 & 0.7873 & 0.4374 & 0.5091 \\
& \textbf{LATTICE} \citep{lattice} & 0.6131 & 0.7558 & 0.5223 & 0.5762 & 0.5306 & 0.6612 & 0.4531 & 0.5036 & 0.5635 & 0.7955 & 0.4110 & 0.5038 \\
& \textbf{MoRec} \citep{morec} & 0.6059 & 0.7552 & 0.5074 & 0.5642 & 0.5267 & 0.6515 & 0.4537 & 0.4970 & 0.5742 & 0.7725 & 0.4277 & 0.4887 \\
& \textbf{BM3} \citep{bm3}  & 0.6222 & 0.7618 & 0.5259 & 0.5788 & 0.5421 & 0.6836 & 0.4749 & 0.5291 & 0.5825 & 0.8155 & 0.4213 & 0.5358 \\
& \textbf{AB-Rec} \citep{ab-rec}  & 0.6194 & 0.7513 & 0.5219 & 0.5787 & 0.5426 & 0.6740 & 0.4645 & 0.5125 & 0.5865 & 0.8094 & 0.4390 & 0.5243 \\
\midrule
\multirow{9}{*}{\textbf{LLM-based}}
& \textbf{TallRec} \citep{tallrec}   & 0.6139 & 0.7324 & 0.5236 & 0.5613 & 0.5243 & 0.6597 & 0.4439 & 0.4958 & 0.5690 & 0.7885 & 0.4174 & 0.4967 \\
& \textbf{LLaRA} \citep{llara}  & 0.6172 & 0.7490 & 0.5201 & 0.5681 & 0.5213 & 0.6799 & 0.4423 & 0.5080 & 0.5650 & 0.7700 & 0.4072 & 0.4854 \\
& \textbf{SPRec} \citep{sprec} & 0.6273 & 0.7598 & 0.5290 & 0.5738 & 0.5473 & 0.6798 & 0.4690 & 0.5138 & 0.5765 & 0.7982 & 0.4250 & 0.5167 \\
& \textbf{RecZero} \citep{reczero} & 0.6595 & 0.7664 & 0.5639 & 0.6067 & 0.6032 & 0.7062 & 0.5166 & 0.5648 & 0.6520 & 0.8205 & 0.5195 & 0.5857 \\
& \textbf{MSRBench} \citep{msrbench} & 0.6272 & 0.7507 & 0.5288 & 0.5743 & 0.5543 & 0.6897 & 0.4739 & 0.5258 & 0.5720 & 0.7820 & 0.4295 & 0.5029 \\
& \textbf{MLLM-MSR} \citep{mllmmsr} & 0.6261 & 0.7563 & 0.5313 & 0.5827 & 0.5447 & 0.6734 & 0.4655 & 0.5181 & 0.5910 & 0.8080 & 0.4429 & 0.5272 \\
& \textbf{MLLMRec-R1} \citep{mllmrec-r1} & 0.7630 & 0.8368 & 0.6524 & 0.6784 & 0.6627 & 0.7906  & 0.5845& 0.6365& 0.7150  & 0.8670& 0.5902& 0.6293 \\
& \cellcolor{clrYellow} \textbf{Qwen3-4B + SFT}    & \cellcolor{clrYellow}0.6104 & \cellcolor{clrYellow}0.7315 &  \cellcolor{clrYellow}0.5228 &  \cellcolor{clrYellow}0.5599 &  \cellcolor{clrYellow}0.5187 & \cellcolor{clrYellow}0.6526 & \cellcolor{clrYellow}0.4339 & \cellcolor{clrYellow}0.4982 & \cellcolor{clrYellow}0.5760 & \cellcolor{clrYellow}0.7510 & \cellcolor{clrYellow}0.4297 & \cellcolor{clrYellow}0.4995 \\
& \cellcolor{clrBlue} \textbf{Qwen3-4B + GRPO}    &  \cellcolor{clrBlue}0.7432 &  \cellcolor{clrBlue}0.8261 &  \cellcolor{clrBlue}0.6333 &  \cellcolor{clrBlue}0.6495 &  \cellcolor{clrBlue}0.6621 &  \cellcolor{clrBlue}0.7764 &  \cellcolor{clrBlue}0.5841 &  \cellcolor{clrBlue}0.6330 &  \cellcolor{clrBlue}0.6690 &  \cellcolor{clrBlue}0.8510 &  \cellcolor{clrBlue}0.5490 &  \cellcolor{clrBlue}0.6198 \\
&  \cellcolor{clrOurs} \textbf{Qwen3-4B + RPSD (Ours)}    &  \cellcolor{clrOurs}\textbf{0.7871} &  \cellcolor{clrOurs}\textbf{0.8716} &  \cellcolor{clrOurs}\textbf{0.6678} &  \cellcolor{clrOurs}\textbf{0.7007} &  \cellcolor{clrOurs}\textbf{0.6836} &  \cellcolor{clrOurs}\textbf{0.7809} &  \cellcolor{clrOurs}\textbf{0.6036} &  \cellcolor{clrOurs}\textbf{0.6417} &  \cellcolor{clrOurs}\textbf{0.8020} &  \cellcolor{clrOurs}\textbf{0.9245} &  \cellcolor{clrOurs}\textbf{0.6952} &  \cellcolor{clrOurs}\textbf{0.7448} \\
\bottomrule
\end{tabular}}
\end{table}
\section{Experiments}
\label{sec:experiments}

Extensive experiments are conducted to answer six questions.
\textbf{RQ1}: How does \method{} perform against state-of-the-art
sequential recommendation methods?
\textbf{RQ2}: Does each design of RPSD contribute to performance improvement?
\textbf{RQ3}: How do the design choices in constructing retrieval-based privileged evidence affect performance?
\textbf{RQ4}: How well does \method{} generalize across domains?
\textbf{RQ5}: How does \method{} behave during training?
\textbf{RQ6}: Does \method{} maintain its performance gains throughout training?


\textbf{Datasets and Metrics.}
We evaluate \method{} on three widely used benchmarks, namely
\textbf{MovieLens}~\citep{movielens}, \textbf{Netflix}~\citep{netflix}, and
\textbf{MicroLens}~\citep{microlens}. Each item is represented by
its ID and title.
Following the standard protocol~\citep{lightgcn}, we adopt hit ratio (HR@$k$)
and normalized discounted cumulative gain (NDCG@$k$, abbreviated as NG@$k$) with
$k\in\{3,5\}$. During
training, the model is optimized to predict the single next item, which provides a
deterministic output for the rule-based reward. At evaluation, rank-based metrics
require a ranked list, so the model instead outputs the $k$ items it considers most
likely. For each test sequence, the ground-truth next item is mixed with $n_c-1$
randomly sampled negatives to form a candidate set, and the Top-3
and Top-5 lists are generated independently for each $k$. The performance gains of
RPSD are statistically significant under a paired $t$-test ($p < 0.05$).
\begin{table}[t]
\centering
\caption{Recommendation performance with a larger candidate set ($n_c=100$).}
\label{tab:more_candidates}
\setlength{\tabcolsep}{10pt}
\renewcommand{\arraystretch}{1.25}
\resizebox{\linewidth}{!}{
\begin{tabular}{l cccc cccc}
\toprule
& \multicolumn{4}{c}{\textbf{MovieLens-1M}} & \multicolumn{4}{c}{\textbf{Netflix}} \\
\cmidrule(lr){2-5}\cmidrule(lr){6-9}
\textbf{Model} & HR@5 & HR@10 & NG@5 & NG@10 & HR@5 & HR@10 & NG@5 & NG@10 \\
\midrule
\textbf{CL4SRec} \citep{cl4srec}  & 0.1427 & 0.1991 & 0.1084 & 0.1198 & 0.1080 & 0.1872 & 0.0671 & 0.0937 \\ \textbf{TallRec} \citep{tallrec}  & 0.1317 & 0.1930 & 0.0930 & 0.1108 & 0.0980 & 0.1790 & 0.0591 &  0.0836\\
\textbf{MLLMRec-R1} \citep{mllmrec-r1}  & 0.2444 & 0.2983 & 0.1806 & 0.1969 & 0.1715 & 0.2560 & 0.1199 & 0.1512 \\
\cellcolor{clrYellow} \textbf{Qwen3-4B + SFT}      & \cellcolor{clrYellow}0.0795 & \cellcolor{clrYellow}0.1134 & \cellcolor{clrYellow}0.0504 & \cellcolor{clrYellow}0.0596 & \cellcolor{clrYellow}0.1155 & \cellcolor{clrYellow}0.1945  & \cellcolor{clrYellow}0.0663 & \cellcolor{clrYellow}0.0910 \\
\cellcolor{clrBlue} \textbf{Qwen3-4B + GRPO}      & \cellcolor{clrBlue}0.2440 & \cellcolor{clrBlue}0.2974 & \cellcolor{clrBlue}0.1802 & \cellcolor{clrBlue}0.1958 & \cellcolor{clrBlue}0.1706 & \cellcolor{clrBlue}0.2541  & \cellcolor{clrBlue}0.1149 & \cellcolor{clrBlue}0.1503 \\
\cellcolor{clrOurs} \textbf{Qwen3-4B + RPSD (Ours)} & \cellcolor{clrOurs}\textbf{0.2740}  & \cellcolor{clrOurs}\textbf{0.3452} &\cellcolor{clrOurs}\textbf{0.1962} & \cellcolor{clrOurs}\textbf{0.2163} & \cellcolor{clrOurs}\textbf{0.3030} & \cellcolor{clrOurs}\textbf{0.3760} & \cellcolor{clrOurs}\textbf{0.2266} & \cellcolor{clrOurs}\textbf{0.2466} \\
\bottomrule
\end{tabular}}
\end{table}
\subsection{Performance Comparison Against SOTA methods (RQ1)}
\label{sec:main_results}

Table~\ref{tab:main_results} reports the comparison under the standard setting
($n_c=10$), from which we draw several observations. First, \method{} performs
strongly against both Non-LLM and LLM-based baselines. Notably, simply replacing
conventional recommenders with an LLM does not consistently yield large gains.
Qwen3-4B + SFT remains comparable to, and in some cases weaker than, strong
Non-LLM baselines. This suggests that the semantic prior of an LLM alone is not
sufficient for accurate recommendation. Clear improvements emerge after
reinforcement learning, indicating that performance depends not
only on understanding item semantics and user preference, but also on aligning
the model with the ranking objective through task-specific feedback.

Second, \method{} consistently outperforms GRPO with the same backbone, showing
that the gain cannot be attributed to increased model capacity. The improvement
instead comes from how the training signal is constructed and assigned. \method{} grounds the teacher in retrieved behavioral
evidence from similar users and uses the resulting teacher--student confidence
gap to redistribute the sequence-level advantage at the token level. This allows
the optimization to focus more on decision-relevant tokens while grounding the
target in cross-user behavioral patterns.

To evaluate \method{} under a more challenging ranking setting, we further enlarge
the candidate set to $n_c=100$ by sampling additional negatives, with results
reported in Table~\ref{tab:more_candidates}. As the candidate pool grows,
distinguishing the target from many competing items becomes substantially harder,
because the model needs to make a more fine-grained ranking decision rather than
separating the target from only a few alternatives. The retrieved evidence enlarges the context that the teacher sees, while token-level credit
concentrates the update on the parts of the response that determine the final
choice. Consequently, \method{} remains effective as the ranking ambiguity
increases.

\begin{table}[t]
\centering
\caption{Ablation studies on the contribution of each proposed module in \method{}.}
\label{tab:ablation on each design}
\setlength{\tabcolsep}{4.5pt}
\resizebox{\linewidth}{!}{
\begin{tabular}{l cccc cccc cccc}
\toprule
\multirow{2}{*}{Variants} & \multicolumn{4}{c}{MovieLens-1M} &
\multicolumn{4}{c}{MicroLens} & \multicolumn{4}{c}{Netflix} \\
\cmidrule(lr){2-5}\cmidrule(lr){6-9}\cmidrule(lr){10-13}
& HR@3 & HR@5 & NG@3 & NG@5 & HR@3 & HR@5 & NG@3 & NG@5 & HR@3 & HR@5 & NG@3 & NG@5 \\
\midrule
GRPO   &   0.7432 &   0.8261 &   0.6333 &   0.6495 &  0.6621 &  0.7764 &  0.5841 &   0.6330 &   0.6690 &  0.8510 &  0.5490 &   0.6198 \\
\method{} w/o OT & 0.7639 & \textbf{0.8716} & 0.6593 & \textbf{0.7007} & 0.6672 & 0.7801 & 0.5881 & 0.6360 & 0.7090 & 0.8660 & 0.5864 & 0.6509 \\
\cellcolor{clrOurs}\textbf{\method{} (Full)}    &  \cellcolor{clrOurs}\textbf{0.7871} &  \cellcolor{clrOurs}\textbf{0.8716} &  \cellcolor{clrOurs}\textbf{0.6678} &  \cellcolor{clrOurs}\textbf{0.7007} &  \cellcolor{clrOurs}\textbf{0.6836} &  \cellcolor{clrOurs}\textbf{0.7809} &  \cellcolor{clrOurs}\textbf{0.6036} &  \cellcolor{clrOurs}\textbf{0.6417} &  \cellcolor{clrOurs}\textbf{0.8020} &  \cellcolor{clrOurs}\textbf{0.9245} &  \cellcolor{clrOurs}\textbf{0.6952} &  \cellcolor{clrOurs}\textbf{0.7448} \\
\bottomrule
\end{tabular}}
\end{table}

\subsection{Ablation Study on the contribution of each proposed module (RQ2)}
In this subsection, we compare three variants to examine the effect of each proposed module. All
variants share the same backbone Qwen3-4B. RPSD w/o OT denotes the model that removes retrieved evidence, keeping only the ground-truth item as the privileged knowledge. RPSD (Full) denotes the full model we propose.
Table~\ref{tab:ablation on each design} shows that RPSD w/o OT improves
recommendation performance even when the teacher receives only the
ground-truth item as privileged information. With the same reward signal as GRPO, the teacher enables the model to reassign token-level advantage, so that optimization focuses more on the tokens that determine the final recommendation. Since the token weights never change the sign of
the advantage, this gain comes purely from redistributing credit within each
response rather than from a stronger or different reward. This confirms that
credit-blind optimization is a real bottleneck of GRPO in recommendation.

The full RPSD matches or outperforms this variant across all metrics, showing
that retrieved evidence provides useful guidance beyond the label itself. When the teacher is conditioned only on the ground-truth label, it knows which item is correct but receives little information about why that item matches the user's history. As a result, the resulting guidance provides only a weak bridge between the observed history and the target recommendation. In contrast, behavioral evidence from similar users exposes concrete preference patterns.

\subsection{Ablation Study on different designs of retrieved evidence (RQ3)}
\label{sec:ablation}

In this subsection, we evaluate five variants to isolate each component.
w/o Re-ranking retrieves evidence by coarse cosine
similarity of mean embeddings, skipping the unbalanced OT re-ranking;
w/o Temporal Cost ($\gamma=0$) drops the temporal alignment
penalty in Eq. (\ref{eq:cost}).
w/o Recency Weighting ($\beta=0$) replaces the recency mass in
Eq. (\ref{eq:recency}) with uniform weights.
w/o Diversity Selection ($\lambda_{\mathrm{div}}=0$) selects
the top-$|\mathcal{S}|$ windows purely by relevance and
w/o Centering \& Clipping uses the raw gap $\Delta_{i,t}$ instead
of $\delta_{i,t}$ in Eq. (\ref{eq:token_weight}).

\begin{table}[t]
\centering
\caption{Ablation studies on different designs of retrieval-based privileged evidence.}
\label{tab:ablation}
\setlength{\tabcolsep}{2.5pt}
\resizebox{\linewidth}{!}{
\begin{tabular}{l cccc cccc cccc}
\toprule
\multirow{2}{*}{Variants} & \multicolumn{4}{c}{MovieLens-1M} &
\multicolumn{4}{c}{MicroLens} & \multicolumn{4}{c}{Netflix} \\
\cmidrule(lr){2-5}\cmidrule(lr){6-9}\cmidrule(lr){10-13}
& HR@3 & HR@5 & NG@3 & NG@5 & HR@3 & HR@5 & NG@3 & NG@5 & HR@3 & HR@5 & NG@3 & NG@5 \\
\midrule
w/o Re-ranking        & 0.7498 & 0.8451 & 0.6398 & 0.6789 & 0.6380 & 0.7629 & 0.5556 & 0.6062 & 0.6950 & 0.8640 & 0.5727 & 0.6404 \\
w/o Temporal Cost        & 0.7481 & 0.8600 & 0.6421 & 0.6885 & 0.6305 & 0.7577 & 0.5517 & 0.6039 & 0.6950 & 0.8545 & 0.5739 & 0.6393 \\
w/o Recency Weighting    & 0.7655 & 0.8600 & 0.6556 & 0.6875 & 0.6358 & 0.7681 & 0.5538 & 0.6094 & 0.6785 & 0.8500 & 0.5563 & 0.6258 \\
w/o Diversity Selection  & 0.7457 & 0.8451 & 0.6414 & 0.6815 & 0.6171 & 0.7487 & 0.5460 & 0.6010 & 0.6810& 0.8575 & 0.5592 & 0.6249 \\
w/o Centering \& Clipping&  0.7639& 0.8563 & 0.6499 & 0.6833 & 0.6432 & 0.7554 & 0.5634 & 0.6094 & 0.6920 & 0.8480 & 0.5662 & 0.6272 \\
\cellcolor{clrOurs}\textbf{Qwen3-4B + RPSD (Ours)}    &  \cellcolor{clrOurs}\textbf{0.7871} &  \cellcolor{clrOurs}\textbf{0.8716} &  \cellcolor{clrOurs}\textbf{0.6678} &  \cellcolor{clrOurs}\textbf{0.7007} &  \cellcolor{clrOurs}\textbf{0.6836} &  \cellcolor{clrOurs}\textbf{0.7809} &  \cellcolor{clrOurs}\textbf{0.6036} &  \cellcolor{clrOurs}\textbf{0.6417} &  \cellcolor{clrOurs}\textbf{0.8020} &  \cellcolor{clrOurs}\textbf{0.9245} &  \cellcolor{clrOurs}\textbf{0.6952} &  \cellcolor{clrOurs}\textbf{0.7448} \\
\bottomrule
\end{tabular}}
\end{table}

Table~\ref{tab:ablation} shows diversity selection is the most consistently important component, with its removal causing the largest degradation on MovieLens-1M and MicroLens. On Netflix, recency weighting plays an equally decisive role, indicating that component importance is somewhat dataset-dependent. Temporal alignment and semantic similarity contribute comparably overall, with neither consistently dominating, suggesting \method{} depends on balancing both cues rather than either alone. Centering and clipping are primarily introduced to stabilize optimization. However, on Netflix, removing them also causes a substantial drop in final recommendation performance. 
Notably, removing OT re-ranking even underperforms removing retrieved evidence altogether (\textit{w/o OT} in Table~\ref{tab:ablation on each design}), suggesting that poorly matched evidence misleads the teacher.

 \begin{figure*}[t]
\includegraphics[width=\textwidth]{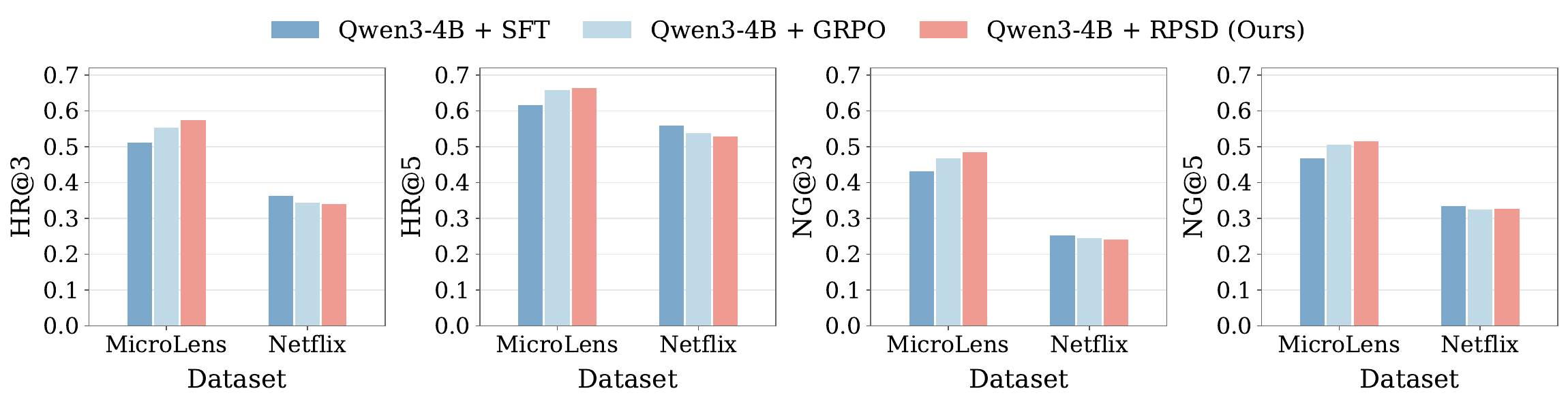}
  \vspace{-0.3in}\caption{\textbf{Cross-domain recommendation performance.} All methods are trained on MovieLens and evaluated on MicroLens and Netflix.}
\label{cross_evaluation}
\end{figure*}

\subsection{Cross-domain Evaluation (RQ4)}

We evaluate cross-domain generalization by training all methods on MovieLens and testing them on MicroLens and Netflix. As shown in Figure~\ref{cross_evaluation}, RPSD achieves the best performance across all metrics on MicroLens, showing that its retrieval-privileged token-level guidance transfers well to this target domain. On Netflix, however, SFT remains slightly stronger than both post-training methods. GRPO shows a similar drop, suggesting that the negative transfer of RPSD is not caused specifically by the privileged teacher. Instead, it may come from source-domain specialization introduced during reinforcement-learning post-training. Overall, these results suggest that the transferability of RPSD depends on how compatible the source and target domains are.

\begin{figure}[t]
\centering
\includegraphics[width=\linewidth]{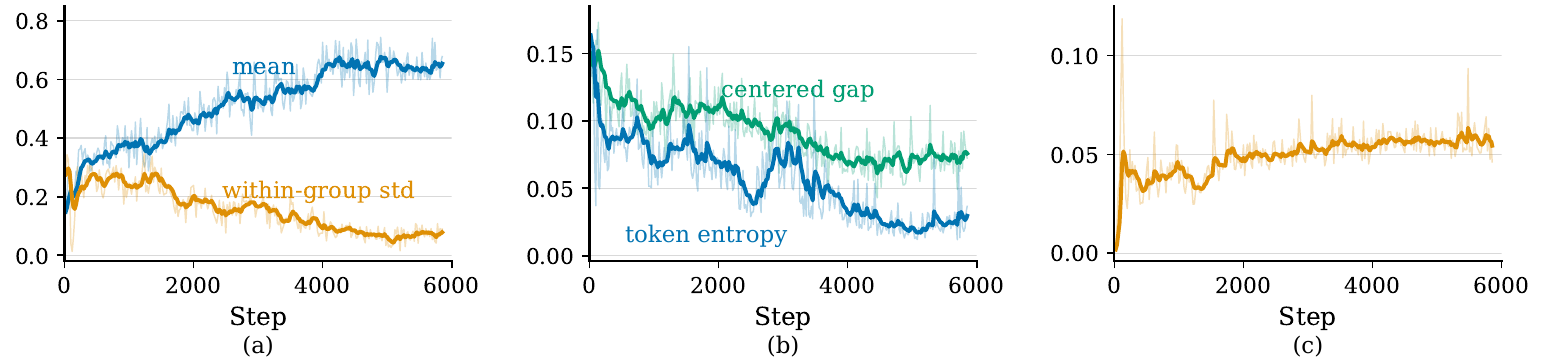}
\vspace{-0.3in}
\caption{\textbf{Training diagnostics of RPSD.} (a) The rule-based reward and the within-group standard deviation of that reward. (b) The centered teacher--student gap $\delta_{i,t}$ and the student's token entropy. (c) The KL divergence to the reference model.}
\label{training diagnostics}
\end{figure}
 \begin{figure*}[t]
\includegraphics[width=\textwidth]{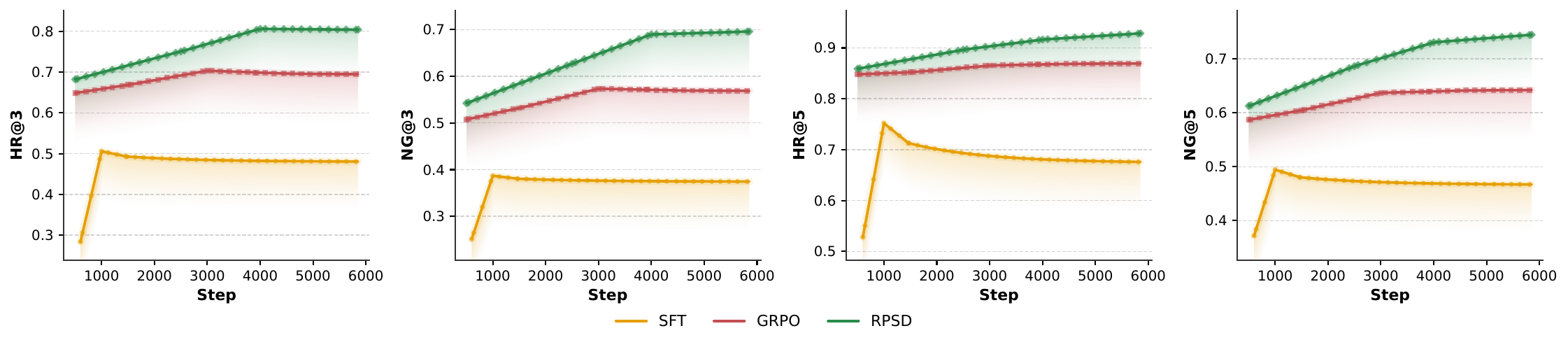}
  \vspace{-0.3in}\caption{\textbf{Performance at different training steps of SFT, GRPO, and RPSD.} RPSD consistently achieves stronger ranking performance and converges to the best final results.}
\label{different steps}
\end{figure*}

\subsection{Training diagnostics of \method{} (RQ5)}
Figure~\ref{training diagnostics} tracks the training dynamics of \method{}. The reward increases steadily over training, and the within-group
standard deviation decreases, showing that the sampled responses become
progressively more consistent. This indicates stable policy optimization. To
measure how confident the student is, we present the token entropy
$\mathcal{H}=-\mathbb{E}_{i,t}\sum_{x\in\mathcal{X}}\pi_{i,t}(x)\log\pi_{i,t}(x)$,
where $\pi_{i,t}(\cdot)=\pi_{\theta}(\cdot\mid x^{S},o_{i,<t})$ is the
student's next-token distribution, $\mathcal{X}$ is the token vocabulary, and
the expectation is taken over all tokens in the sampled group. A lower
$\mathcal{H}$ means the student places its probability on fewer tokens. As
shown in Figure~\ref{training diagnostics} (b), $\mathcal{H}$ falls quickly during
training, which means the student gradually commits to the few tokens. The
teacher--student gap also decreases, so privileged guidance is most useful in
the early stage, when the student is still uncertain. As training continues,
the student becomes more confident. Interestingly, the gap stays clearly above zero even after
$\mathcal{H}$ has nearly flattened. A confident student is therefore not yet
fully aligned with the teacher, and the retrieved evidence keeps providing
token-level signal late in training rather than fading away with the
uncertainty. The KL divergence to the reference model rises early and then stabilizes.

\subsection{Performance Across Training Steps (RQ6)}
Figure~\ref{different steps} evaluates the performance of SFT, GRPO, and RPSD
at different training steps. RPSD outperforms GRPO from the first evaluated checkpoint and maintains this advantage throughout training. This shows that its benefits appear early and persist until convergence. The larger gains in NG than in HR suggest that RPSD not only retrieves the target item more often but also ranks it higher. These results support the effectiveness of using collaborative evidence to assign token-level credit. In contrast, SFT shows little improvement over training. Since all three methods share the same backbone, the stronger performance of RPSD can be attributed to our designs.

\section{Conclusion}
\label{sec:conclusion}

In this paper, we propose \method{} to address credit-blind optimization and collaborative-blind supervision. \method{} enriches the OPSD teacher with cross-user behavioral evidence retrieved by temporally aware optimal transport and diversity-aware selection, enabling token-level supervision that captures both target correctness and collaborative behavioral patterns. Extensive experiments on multiple benchmarks show that \method{} consistently outperforms state-of-the-art methods.

\section*{AI use statement}
We use generative AI tools to polish the writing of this manuscript. All AI-assisted content is reviewed by the authors, who take full responsibility for the final content.

\bibliography{iclr2027_conference}
\bibliographystyle{iclr2027_conference}

\appendix

\section{Appendix Overview}

The appendix is organized as follows.

\begin{itemize}
    \item \textbf{Algorithm Details.} We provide the algorithm details of the complete RPSD training procedure, the evidence retrieval pipeline, and the unbalanced optimal transport solver.
    \item \textbf{Implementation Details.} We report the model configuration, training settings, retrieval setup, and hyper-parameters used to reproduce RPSD.
    \item \textbf{Dataset Details.} We describe the three benchmark datasets used in the main paper. Then we report their statistics and how we construct the training, validation, and test sets.
    \item \textbf{Evaluation Protocol.} We specify the candidate construction, evaluation metrics, ranking procedure, and repeated-run protocol.
    \item \textbf{Baseline Details.} We describe the Non-LLM and LLM-based recommenders used for comparison, together with their evaluation settings.
    \item \textbf{Training Dynamics of RPSD.} We analyze how RPSD behaves during optimization across datasets, in terms of teacher--student guidance, reward, entropy, and policy stability.
    \item \textbf{Training Dynamics of RPSD w/o OT.} We study the training dynamics of RPSD w/o OT in order to isolate the role of behavioral evidence retrieved using OT.
    \item \textbf{Comparison with Vanilla OPSD.} We compare RPSD with vanilla OPSD, which matches the student to a label-conditioned teacher distribution. We show that keeping the reward-grounded objective while reweighting token-level credit leads to consistent gains.
    \item \textbf{Response-Level RPSD versus Token-Level RPSD.} We apply the same teacher-derived weights at the response level rather than the token level. This isolates the effect of credit granularity and shows that the improvement of RPSD comes from token-specific credit.
    \item \textbf{Case Study of Student and Teacher Prompts.} We illustrate the student and teacher prompts, including the retrieved evidence block, using a training example from MovieLens-1M.

\end{itemize}
\section{Algorithm Details}
\label{app:algorithm}

The full procedure is given in Algorithm~\ref{alg:rpsd}. Evidence retrieval is
detailed in Algorithm~\ref{alg:retrieval}, and the optimal-transport solver at
its core is given in Algorithm~\ref{alg:sinkhorn}.

During training, the student samples a group of responses from the student
prompt $x^{S}$. Each response receives a rule-based reward, and GRPO turns these
rewards into a single group-normalized advantage $\widehat A_i$. A teacher that additionally sees the retrieved evidence $E_u$
then rescores the same tokens under stop-gradient. From this, we compute a
confidence gap $\Delta_{i,t}$, which we center within each completion and clip
into $\delta_{i,t}$.
We map $\delta_{i,t}$ to an advantage-aware weight $w_{i,t}$ and use it to refine
$\widehat A_i$ into a token-level advantage $\widetilde A_{i,t}$. As a result,
tokens supported by the evidence are reinforced more strongly when the response
is correct, and penalized more strongly when it is wrong.
Finally, we optimize the clipped GRPO objective with $\widetilde A_{i,t}$ to
update the policy. At inference, the model conditions only on $x^{S}$. Therefore, the
evidence shapes token credit during training and is never used at test time.

\begin{algorithm}[t]
\caption{RPSD training}
\label{alg:rpsd}
\begin{algorithmic}[1]
\REQUIRE training set, policy $\pi_\theta$ initialized from the SFT model,
group size $G$, guidance strength $\lambda$, weight bound $\epsilon_w$,
stability threshold $c_\delta$, clip range $\epsilon$, KL weight $\beta_{\mathrm{KL}}$
\ENSURE trained policy $\pi_\theta$
\STATE build the evidence index $\mathcal{W}$
\FOR{each training step}
  \STATE sample a user with history $H_u$, candidate set $\mathcal{C}_u$, and ground-truth next item $v^u_{t+1}$
  \STATE $x^{S} \gets \textsc{Prompt}_{S}(H_u,\mathcal{C}_u)$ 
  \STATE $E_u \gets \textsc{RetrieveEvidence}(H_u,\mathcal{W})$ 
  \STATE $x^{T} \gets x^{S}\oplus E_u\oplus v^{u}_{t+1}$ 
  \STATE sample a group $\{o_i\}_{i=1}^{G}\sim \pi_\theta(\cdot\mid x^{S})$
  \STATE $r_i \gets +1$ if $o_i$ is valid and names the target, $0$ if valid but wrong, $-1$ if invalid
  \STATE $\widehat A_i \gets \big(r_i-\operatorname{mean}(r)\big)/\operatorname{std}(r)$
  \FOR{each token $o_{i,t}$}
    \STATE $\Delta_{i,t} \gets \log \pi_{\operatorname{sg}(\theta)}(o_{i,t}\mid x^{T},o_{i,<t}) - \log \pi_{\operatorname{sg}(\theta)}(o_{i,t}\mid x^{S},o_{i,<t})$
    \STATE $\delta_{i,t} \gets \operatorname{clip}\big(\Delta_{i,t}-\tfrac{1}{T_i}\textstyle\sum_{t'}\Delta_{i,t'},\,-c_\delta,\,c_\delta\big)$
    \STATE $w_{i,t} \gets \operatorname{clip}\big(\exp[\operatorname{sign}(\widehat A_i)\,\delta_{i,t}],\,1-\epsilon_w,\,1+\epsilon_w\big)$
    \STATE $\widetilde A_{i,t} \gets \widehat A_i\big[(1-\lambda)+\lambda\,w_{i,t}\big]$
  \ENDFOR
  \STATE update $\theta$ by minimizing the RPSD objective in Eq. (\ref{eq:opsd_objective})
\ENDFOR
\RETURN $\pi_\theta$
\end{algorithmic}
\end{algorithm}

Evidence retrieval is summarized in Algorithm~\ref{alg:retrieval}. Each item is
encoded from its title text alone, using the mean-pooled hidden state of the SFT
model. We reduce these vectors with PCA, and then index every length-$K$ window
together with the item.
Given a query history, retrieval proceeds in two stages. We first perform a coarse retrieval of
candidate windows by cosine similarity of recency-weighted mean embeddings,
excluding the query user's own windows so that no target leaks. We then re-rank
the coarsely retrieved candidate windows with unbalanced entropic optimal transport, which aligns the two
item sequences under a cost that combines semantic dissimilarity with a temporal
penalty. Finally, a greedy criterion selects a diverse subset, which is rendered
into the evidence block $E_u$.

\begin{algorithm}[t]
\caption{Retrieve Evidence for a query history $H_u$}
\label{alg:retrieval}
\begin{algorithmic}[1]
\REQUIRE training sequences $\{S_u\}$, item titles $\{t_v\}$, SFT model $f$,
window length $K$, recency rate $\beta$, embedding dimension $d$, recall size
$N_0$, OT parameters $\varepsilon,\rho$, cost weight $\gamma$, temperature $\tau$,
diversity weight $\lambda_{\mathrm{div}}$, evidence count $m$
\ENSURE evidence block $E_u$
\STATE $e_v \gets \operatorname{meanpool}(f(t_v))$ for every item, reduce all $\{e_v\}$ to $d$ dimensions by PCA, and $\ell_2$-normalize them
\STATE $a_p \gets \frac{\exp[-\beta(K-1-p)]}{\sum_{r=0}^{K-1}\exp[-\beta(K-1-r)]}$ for $p=0,\dots,K-1$; \quad $a\gets(a_0,\dots,a_{K-1})$, \quad $b\gets a$
\STATE build $\mathcal{W} \gets \{(h_j,v_j^{+},u_j,\bar e(h_j))\}$ over all length-$K$ windows $h_j$ of the training sequences that are followed by an item $v_j^{+}$, excluding the last window of each sequence, where $\bar e(h_j)=\frac{\sum_p a_p e_{v'_{j,p}}}{\|\sum_p a_p e_{v'_{j,p}}\|_2}$
\IF{$|H_u|<K$}
  \RETURN $E_u$ stating that no similar trajectory is retrieved
\ENDIF
\STATE $H_u^{K} \gets$ the $K$ most recent items of $H_u$; \quad $\bar e(H_u^{K}) \gets$ recency-weighted mean by the same rule
\STATE $\mathcal{R} \gets$ top-$N_0$ windows by cosine $\langle \bar e(H_u^{K}),\bar e(h_j)\rangle$, excluding windows with $u_j=u$ or $h_j=H_u^{K}$
\FOR{each $h_j\in\mathcal{R}$}
  \STATE $C_{pq} \gets 1-\langle e_{v_p},e_{v'_{j,q}}\rangle+\gamma\,\big|\tfrac{p}{K-1}-\tfrac{q}{K-1}\big|$, where $v_p$ is the $p$-th item of $H_u^{K}$
  \STATE $(P^\ast,c_j) \gets$ transport plan and mass-normalized cost for $(a,b,C)$
  \STATE $s_j \gets \exp(-c_j/\tau)$ \COMMENT{$c_j=\langle P^\ast,C\rangle/\|P^\ast\|_1$}
\ENDFOR
\STATE $\mathcal{S} \gets \emptyset$
\WHILE{$|\mathcal{S}|<\min(m,|\mathcal{R}|)$}
  \STATE $j^\ast \gets \arg\max_{j\in\mathcal{R}\setminus\mathcal{S}}\big[s_j-\lambda_{\mathrm{div}}\max_{k\in\mathcal{S}}\langle \bar e(h_j),\bar e(h_k)\rangle\big]$ \COMMENT{inner max $:=0$ if $\mathcal{S}=\emptyset$}
  \STATE $\mathcal{S}\gets\mathcal{S}\cup\{j^\ast\}$
\ENDWHILE
\STATE $E_u \gets$ render each $h_j\in\mathcal{S}$ as one line listing its $K$ items, the item $v_j^{+}$ that user interacted with next, and its similarity $s_j$
\RETURN $E_u$
\end{algorithmic}
\end{algorithm}

\begin{algorithm}[t]
\caption{Unbalanced entropic OT (mass-normalized cost)}
\label{alg:sinkhorn}
\begin{algorithmic}[1]
\REQUIRE query mass $\mathbf{a}\in\mathbb{R}^{K}$, window mass $\mathbf{b}\in\mathbb{R}^{K}$, ground cost $\mathbf{C}\in\mathbb{R}^{K\times K}$, regularization $\varepsilon$, relaxation $\rho$, maximum iterations $T$
\ENSURE transport plan $\mathbf{P}$ and mass-normalized cost $c$
\STATE $\mathbf{K}\gets\exp(-\mathbf{C}/\varepsilon)$
\STATE $\mathbf{u}\gets\mathbf{1}_K,\quad \mathbf{v}\gets\mathbf{1}_K$
\STATE $f\gets\rho/(\rho+\varepsilon)$ \COMMENT{$f\to 1$ recovers balanced OT}
\FOR{$t=1$ \TO $T$}
  \STATE $\mathbf{u}\gets\big(\mathbf{a}\,/\,(\mathbf{K}\mathbf{v})\big)^{f}$
  \STATE $\mathbf{v}\gets\big(\mathbf{b}\,/\,(\mathbf{K}^{\top}\mathbf{u})\big)^{f}$
  \STATE \textbf{break} if $\mathbf{u}$ and $\mathbf{v}$ have converged
\ENDFOR
\STATE $\mathbf{P}\gets\operatorname{diag}(\mathbf{u})\,\mathbf{K}\,\operatorname{diag}(\mathbf{v})$
\STATE $c\gets\langle\mathbf{P},\mathbf{C}\rangle/\|\mathbf{P}\|_1$
\RETURN $(\mathbf{P},c)$
\end{algorithmic}
\end{algorithm}

The re-ranking step of Algorithm~\ref{alg:retrieval} uses the unbalanced entropic optimal-transport solver in Algorithm~\ref{alg:sinkhorn}. Given the query mass $\mathbf{a}$ and window mass $\mathbf{b}$ from the recency-weighted rule, together with a ground cost $\mathbf{C}$, the solver returns the transport plan and its mass-normalized cost, which we map to a similarity. The cost between query position $p$ and window position $q$ is $C_{pq} = (1 - \langle e_p, e'_q\rangle) + \gamma \cdot |p/(K-1) - q/(K-1)|$, where the first term measures semantic dissimilarity between item-title embeddings and the second penalizes temporally inconsistent alignments. The marginal relaxation controlled by $\rho$ lets some mass stay unmatched, so a query and a window that share only part of their interests need not be fully aligned. As a special case, setting $\mathbf{C}$ to an identity kernel reduces the solver to recency-weighted exact matching.

\section{Implementation Details}
\label{sec:implementation_details}

We initialize the policy from an SFT-trained Qwen3-4B checkpoint and optimize it
with LoRA. We set the rank to $32$, the scaling factor to $64$, and the dropout
rate to $0.05$, and we apply adapters to all linear layers.
We split the sequence instances into training, validation, and test sets with a
ratio of 7:1:2. Each instance is built from up to nine recent interactions. We
use the final item as the prediction target and the preceding items as the input
history, so each instance has at most eight history items. We apply the same
construction to all three splits.
For each instance, we build a candidate set of ten items: the target item and
nine negatives. The negatives are sampled at random from the user's
non-interacted items. The student sees only the input history and the candidate
set, and it is not told which candidate is the target.

During training, the privileged teacher additionally receives behavioral evidence
retrieved from other similar users. We first retrieve candidate history windows
using 128-dimensional item embeddings, ranked by recency-weighted cosine
similarity. We then re-rank them with temporally aware unbalanced OT. Finally, we apply MMR-based diversity selection. Specifically, we greedily pick three
evidence windows, balancing their relevance to the query history against their
similarity to windows already selected. For each selected window, we include the
window itself and its subsequent interaction in the teacher's evidence block. This
evidence is used only by the teacher during training, and is not needed at
inference time.

For each training prompt, we sample eight completions from the student policy. A completion
receives reward $1$ if it follows the required format and predicts the correct
item, $0$ if it is well-formed but incorrect, and $-1$ if it violates the output
format.
We then re-evaluate the same policy under the privileged teacher context with
gradients disabled. Here the teacher only scores the student's completion and it does
not generate a separate response. We compute the teacher--student log-probability
difference for each token and center it within each completion.
We train for three epochs and the learning rate is $8\times10^{-6}$. The
per-device batch size is $4$, and we use $8$ gradient-accumulation steps. We set
the KL coefficient to $0.1$ and save a checkpoint every $500$ optimization steps.

\section{Dataset Details}
\label{app:datasets}
We evaluate our model on three public recommendation datasets. They span the
short-video and movie domains, and they differ in interaction density by more than
two orders of magnitude. A statistical overview is given in
Table~\ref{tab:datasets}. MicroLens and Netflix are large and sparse, which means that each user
leaves only a few interactions on these datasets, so a model should generalize from very short
histories. MovieLens is small but extremely dense, which means that long user sequences on this dataset make item
patterns easy to observe. This spread lets us test whether retrieved behavioral
evidence still helps when a user's own history is too short to be informative,
which is exactly the sparse regime our method targets.

\textbf{MicroLens}~\citep{microlens} is a short-video recommendation dataset. Each
item carries a title, a cover image, and video information. Here, we use the title
and the identifier as the item representation. It is the largest among the three datasets,
with $25{,}411$ users and $41{,}081$ items. It is also highly sparse, at a density
of $2.1\times10^{-4}$, so each user interacts with only $8.79$ items on average.
The large catalog and short histories make it a challenging test of generalization.

\textbf{Netflix}~\citep{netflix} is a user-based movie recommendation dataset from
Kaggle, released by LLMRec. Each item comes with a title and a poster image. It has
the shortest histories. Each user interacts with only $5.23$ items on
average, at a density of $3.0\times10^{-4}$. Because a single history reveals so
little, Netflix is the dataset on which evidence from behaviorally similar users should matter
most.

\textbf{MovieLens}~\citep{movielens} is a classic movie recommendation dataset in
which each movie provides a title, a release year, and a poster image. Unlike the
other two datasets, it has a small catalog but is extremely dense, with a density
of $4.2\times10^{-2}$ and an average sequence length of $165.56$. Each user
therefore has a long, information-rich history, which is exactly the setting where
standard sequential models already perform well. This makes MovieLens suitable for our approach. It lets us test whether retrieved evidence still helps even
when a user's own history is already informative enough on its own.

We split the sequence instances into training, validation, and
test sets with a ratio of 7:1:2. Within each instance, we retain
up to the nine most recent interactions, use the final item as
the prediction target, and use the preceding items as the
input history. This instance construction is applied consistently
across the training, validation, and test splits.
\begin{table}[t]
\centering
\caption{Statistics of the experimental datasets.}
\label{tab:datasets}
\begin{tabular}{lcccc}
\toprule
\textbf{Dataset} & \textbf{\#Users} & \textbf{\#Items} & \textbf{\#Interactions} & \textbf{Density} \\
\midrule
MicroLens & 25{,}411 & 41{,}081 & 223{,}263   & $2.1\times10^{-4}$ \\
Netflix   & 13{,}187 & 17{,}366 & 68{,}933    & $3.0\times10^{-4}$ \\
MovieLens & 6{,}040  & 3{,}952  & 1{,}000{,}209 & $4.2\times10^{-2}$ \\
\bottomrule
\end{tabular}
\end{table}

\section{Evaluation Protocol}

For each test instance, we construct a candidate set by combining
the ground-truth next item with $n_c-1$ negatives sampled from
the user's non-interacted items. We use $n_c=10$ by default and
additionally evaluate $n_c=100$ to examine performance with a
larger candidate set. All methods are evaluated using the same
data splits and candidate construction protocol.

We evaluate recommendation performance using hit ratio (HR@$k$) and normalized
discounted cumulative gain (NDCG@$k$). During training, the model is optimized to
predict the single next item, since the rule-based reward requires a deterministic,
verifiable output. For evaluation, rank-based metrics instead require a ranked list,
so we prompt the model to output the $k$ items it considers most likely, forming a
ranked list $\mathcal{L}_i^{(k)}$ for each test instance. The Top-3 and Top-5 lists
are generated independently rather than obtained by truncating a single shared
ranking, and both HR@$k$ and NDCG@$k$ are computed from $\mathcal{L}_i^{(k)}$ at the
corresponding threshold $k$. Since each instance has exactly one ground-truth
target, HR@$k$ measures whether the target appears within the top $k$, and NDCG@$k$
additionally rewards ranking it higher.

Let $\mathcal{T}$ denote the test set. For each instance $i$, let
$\operatorname{rank}_i^{(k)}$ be the one-based position of the ground-truth item in
the list $\mathcal{L}_i^{(k)}$. If the target does not appear in the list, we set
$\operatorname{rank}_i^{(k)}=+\infty$.

HR@$k$ is the fraction of test instances whose target appears within the top $k$
positions:
\begin{equation}
    \mathrm{HR@}k
    =
    \frac{1}{|\mathcal{T}|}
    \sum_{i \in \mathcal{T}}
    \mathbb{I}\!\left[
        \operatorname{rank}_i^{(k)} \leq k
    \right],
\end{equation}
where $\mathbb{I}[\cdot]$ is the indicator function.

NDCG@$k$ further rewards ranking the target higher. Each instance has exactly one
target, and we compute
\begin{equation}
    \mathrm{NDCG@}k
    =
    \frac{1}{|\mathcal{T}|}
    \sum_{i \in \mathcal{T}} g_i^{(k)},
    \qquad
    g_i^{(k)}
    =
    \begin{cases}
        \displaystyle
        \frac{1}{
            \log_2\!\left(
                \operatorname{rank}_i^{(k)}+1
            \right)
        },
        & \operatorname{rank}_i^{(k)} \leq k, \\[6pt]
        0,
        & \text{otherwise}.
    \end{cases}
\end{equation}

We report HR@$k$ and NDCG@$k$ for $k \in \{3,5\}$ under the default candidate
setting, and for $k \in \{5,10\}$ under the larger one. For both metrics, higher is better, and all results are averaged
over five independent runs.

\section{Baseline Details}
\label{app:baselines}

We compare RPSD against two families of sequential recommenders. All baselines are trained and evaluated on the same data splits, the same candidate construction, and the same $n_c$ as RPSD, so that differences reflect the method rather than the protocol.

\textbf{Non-LLM models.}
GRU4Rec~\citep{gru4rec} encodes an interaction session with a gated recurrent
network and predicts the next click from its final hidden state. SASRec~\citep{sasrec}
replaces recurrence with a causal self-attention encoder that adaptively weighs
earlier interactions. LightGCN~\citep{lightgcn} propagates user and item embeddings
over the interaction graph, dropping feature transformation and nonlinearity for a
lightweight collaborative signal. CL4SRec~\citep{cl4srec} adds a contrastive
objective over perturbed sequences to a self-attentive backbone, strengthening its
representations under sparse supervision. These models rely on identifier embeddings
alone, and we use them to measure the effectiveness of purely collaborative signals. LATTICE~\citep{lattice} builds a latent item-to-item graph from modality
features for adding content structure to collaborative filtering. MoRec~\citep{morec}
studies whether modality-derived item representations can replace identifier
embeddings. BM3~\citep{bm3} learns multimodal representations with a self-supervised
bootstrapping objective that needs no negative sampling. AB-Rec~\citep{ab-rec} aligns
and balances identifier and modality representations so that neither dominates the
fused embedding.

\textbf{LLM-based recommenders.}
TallRec~\citep{tallrec} tunes an LLM on recommendation prompts so that
it picks the next item from a candidate list. LLaRA~\citep{llara} connects
collaborative embeddings to a language backbone through a projection, letting
sequential signals enter the prompt. SPRec~\citep{sprec} uses a self-play preference
scheme to reduce popularity bias. MSRBench~\citep{msrbench} studies how large models
can be integrated into sequential recommendation under different strategies, and
MLLM-MSR~\citep{mllmmsr} fine-tunes a large model with an item-summarization stage
that feeds a sequence model. RecZero~\citep{reczero} post-trains an LLM with GRPO
over structured reasoning steps and a graded rule-based reward.

Our own variants, Qwen3 with SFT, Qwen3 with GRPO, and Qwen3 with RPSD, share the identical backbone, prompt, sampling, and reward, and differ only in the method design, so the comparison among them attributes any gain to proposed modules.

\begin{figure}[t]
\centering
\includegraphics[width=\linewidth]{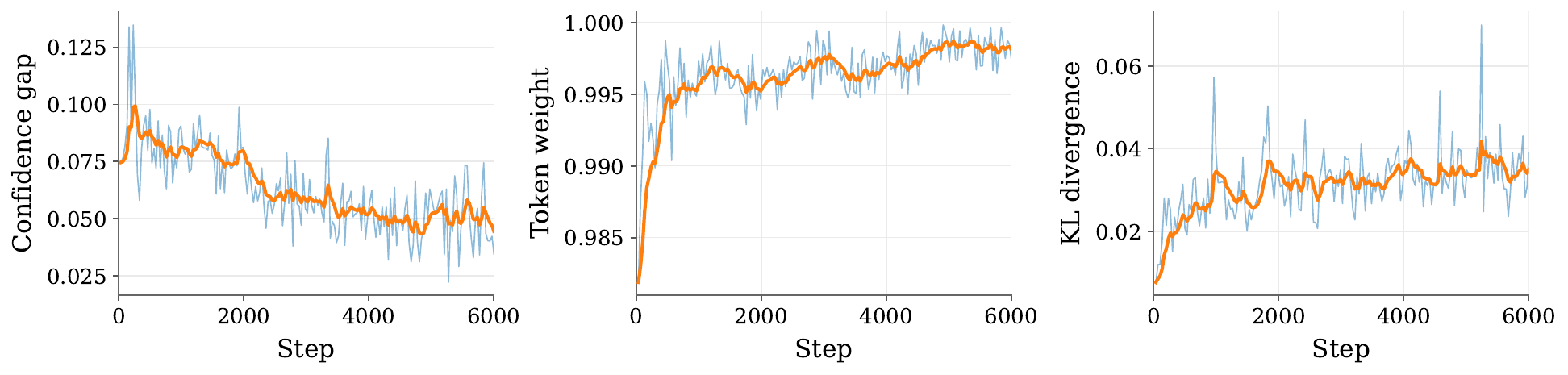}
\caption{\textbf{RPSD training diagnostics on MovieLens.} We track three
quantities that are internal to RPSD.
\textbf{(left)} The teacher-student confidence gap $\delta_{i,t}$ averaged over completion tokens, which measures how much the teacher model differs from the student model.
\textbf{(middle)} The advantage-aware token weight $w_{i,t}$, which sets how strongly the shared advantage is
reshaped. \textbf{(right)} The KL
divergence to the reference policy. Blue curves are the raw logged values and
orange curves are moving averages.}
\label{fig:movielens_diag_a}
\end{figure}

\begin{figure}[t]
\centering
\includegraphics[width=\linewidth]{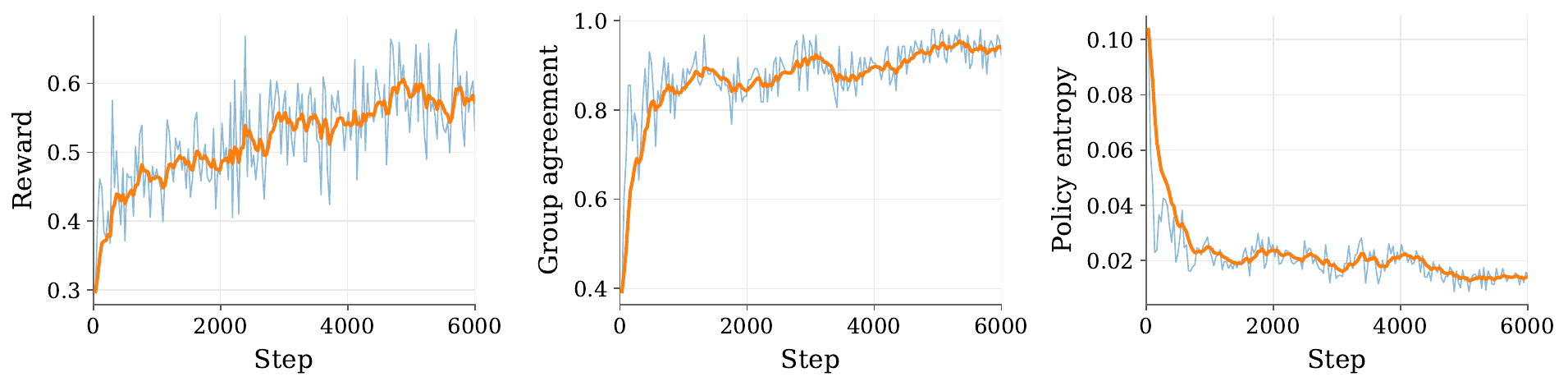}
\caption{\textbf{Training stability of RPSD on MovieLens.} \textbf{(left)}
Rule-based reward. \textbf{(middle)} Group agreement, i.e., the fraction of
sampled groups whose $G$ responses all receive the same reward. Such groups
have zero reward variance and provide no advantage signal. \textbf{(right)}
Policy entropy over completion tokens. Blue curves are the raw logged values
and orange curves are moving averages.}
\label{fig:movielens_diag_b}
\end{figure}

\begin{figure}[t]
\centering
\includegraphics[width=\linewidth]{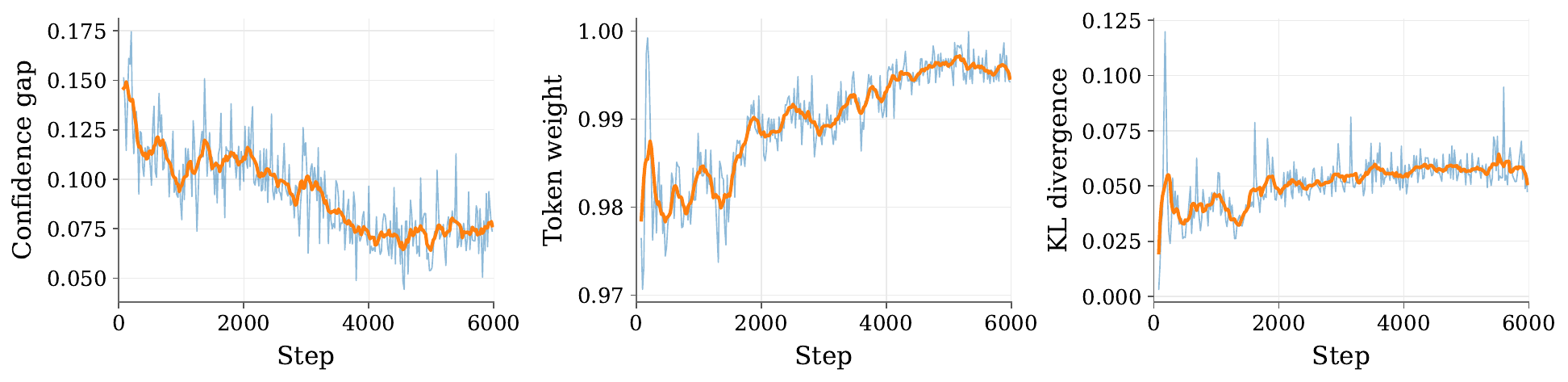}
\caption{\textbf{RPSD training diagnostics on Netflix.} We track three
quantities that are internal to RPSD.
\textbf{(left)} The teacher-student confidence gap $\delta_{i,t}$ averaged over completion tokens, which measures how much the teacher model differs from the student model.
\textbf{(middle)} The advantage-aware token weight $w_{i,t}$, which sets how strongly the shared advantage is
reshaped. \textbf{(right)} The KL
divergence to the reference policy. Blue curves are the raw logged values and
orange curves are moving averages.}
\label{fig:netflix_diag_a}
\end{figure}

\begin{figure}[t]
\centering
\includegraphics[width=\linewidth]{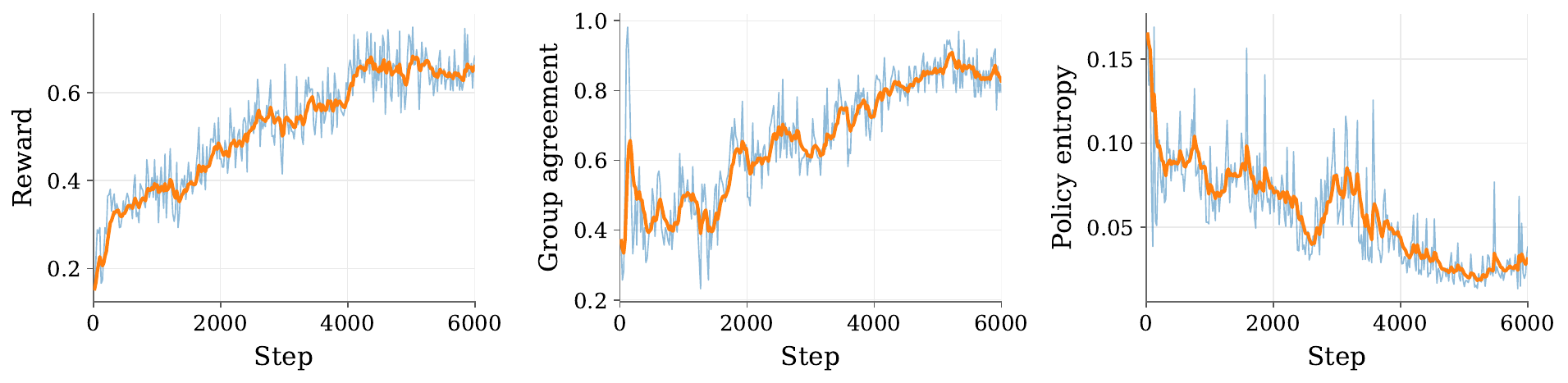}
\caption{\textbf{Training stability of RPSD on Netflix.} \textbf{(left)}
Rule-based reward. \textbf{(middle)} Group agreement, i.e., the fraction of
sampled groups whose $G$ responses all receive the same reward. Such groups
have zero reward variance and provide no advantage signal. \textbf{(right)}
Policy entropy over completion tokens. Blue curves are the raw logged values
and orange curves are moving averages.}
\label{fig:netflix_diag_b}
\end{figure}

\begin{figure}[t]
\centering
\includegraphics[width=\linewidth]{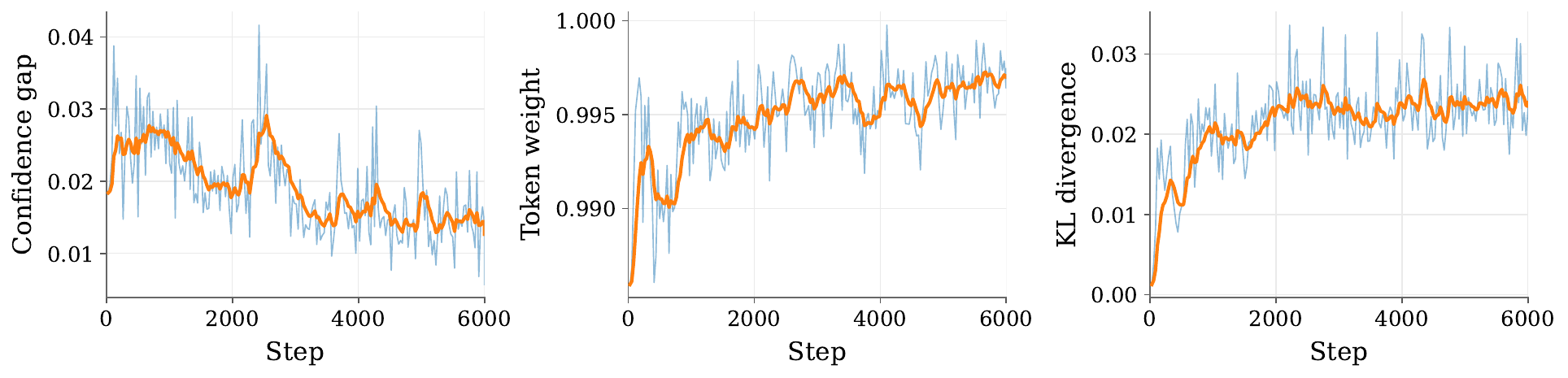}
\caption{\textbf{RPSD training diagnostics on MicroLens.} We track three
quantities that are internal to RPSD.
\textbf{(left)} The teacher-student confidence gap $\delta_{i,t}$ averaged over completion tokens, which measures how much the teacher model differs from the student model.
\textbf{(middle)} The advantage-aware token weight $w_{i,t}$, which sets how strongly the shared advantage is
reshaped. \textbf{(right)} The KL
divergence to the reference policy. Blue curves are the raw logged values and
orange curves are moving averages.}
\label{fig:microlens_diag_a}
\end{figure}

\begin{figure}[t]
\centering
\includegraphics[width=\linewidth]{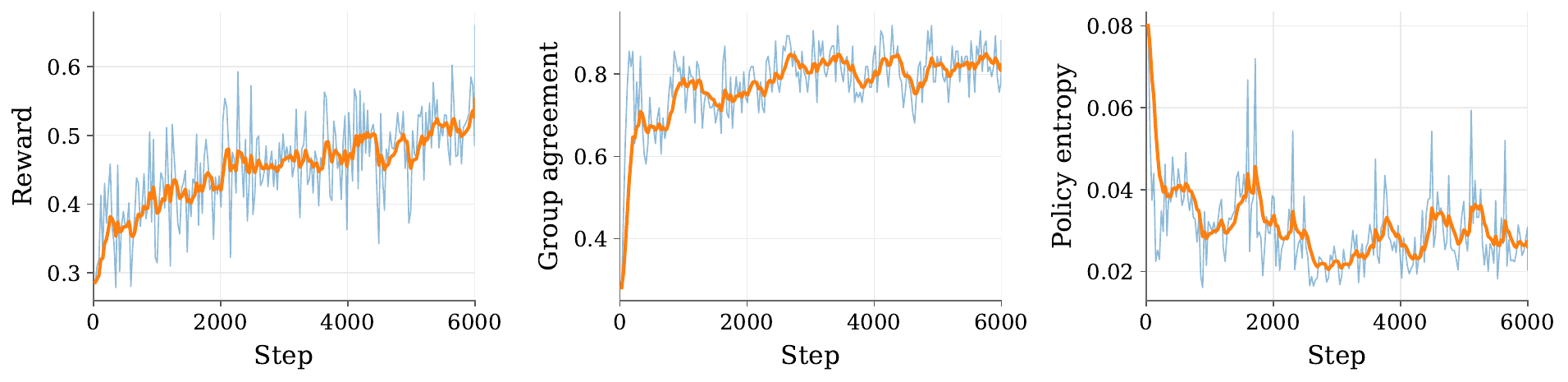}
\caption{\textbf{Training stability of RPSD on MicroLens.} \textbf{(left)}
Rule-based reward. \textbf{(middle)} Group agreement, i.e., the fraction of
sampled groups whose $G$ responses all receive the same reward. Such groups
have zero reward variance and provide no advantage signal. \textbf{(right)}
Policy entropy over completion tokens. Blue curves are the raw logged values
and orange curves are moving averages.}
\label{fig:microlens_diag_b}
\end{figure}

\section{Training Diagnostics of RPSD}

In this section, we track how RPSD optimizes on the three datasets. The
diagnostics are shown in Figures~\ref{fig:movielens_diag_a} - \ref{fig:microlens_diag_b}. Across all of
them, the rule-based reward rises steadily and does not collapse, and the policy
entropy goes down over time. This means the model becomes more confident and
gradually commits to the tokens that identify the target item. The KL divergence to
the reference model rises only at the beginning and then stays flat, so the training
stays close to the reference rather than drifting away from it. Together, these
curves show that RPSD trains in a stable way.

We next look at the two quantities that are specific to RPSD. The teacher--student
confidence gap $\delta_{i,t}$ is largest early in training and becomes smaller as the
policy converges. This shows that the retrieved evidence helps most when the student
is still uncertain, and its effect fades once the student can already score the
decisive tokens on its own. This pattern is consistent across the three datasets, and
the main difference is how far the gap narrows, which reflects how much privileged knowledge already reveals.

The token weight $w_{i,t}$ moves toward one as training goes on. This does not mean
the teacher stops mattering; it follows directly from the shrinking confidence gap,
since $w_{i,t}$ is computed from $\delta_{i,t}$ and approaches one exactly when
$\delta_{i,t}$ becomes small. In other words, RPSD applies a strong correction early,
when the student needs it, and a gentle one later, when the student is already
confident. This happens on its own, without any manually designed annealing schedule,
which makes the token-level guidance easy to use in practice.

The group agreement, defined as the fraction of groups whose $G$ responses receive the same reward, increases steadily during training. This trend is consistent with the behavior of the token weight $w_{i,t}$. As the policy becomes more confident, responses within the same group are more likely to receive the same reward, leading to higher group agreement. At the same time, the teacher--student gap $\delta_{i,t}$ decreases, causing $w_{i,t}$ to increase. Together, these trends indicate that both the group-level reward variation and the teacher's token-level correction gradually weaken as training converges.


\section{Training Dynamics of RPSD w/o OT}
\begin{figure}[t]
\centering
\includegraphics[width=\textwidth]{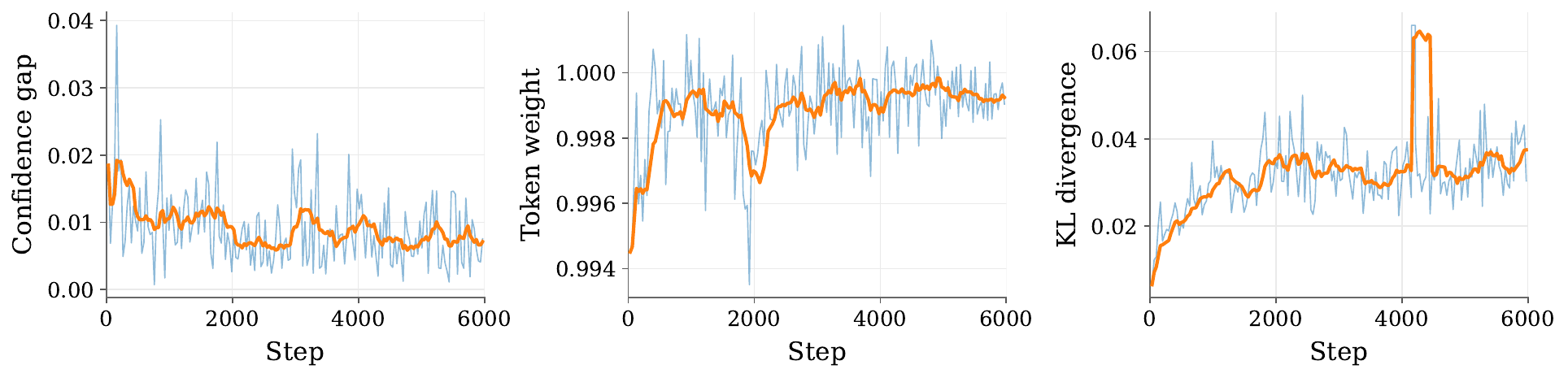}
\caption{\textbf{Training diagnostics of RPSD w/o OT on MovieLens-1M.} The
teacher is privileged with the ground-truth item alone. \textbf{(left)} The
teacher--student confidence gap averaged over completion tokens.
\textbf{(middle)} The advantage-aware token weight. \textbf{(right)}
The KL divergence to the reference policy. Blue curves are the raw
logged values and orange curves are moving averages.}
\label{w/o OT first}
\end{figure}
 
\begin{figure}[t]
\centering
\includegraphics[width=\textwidth]{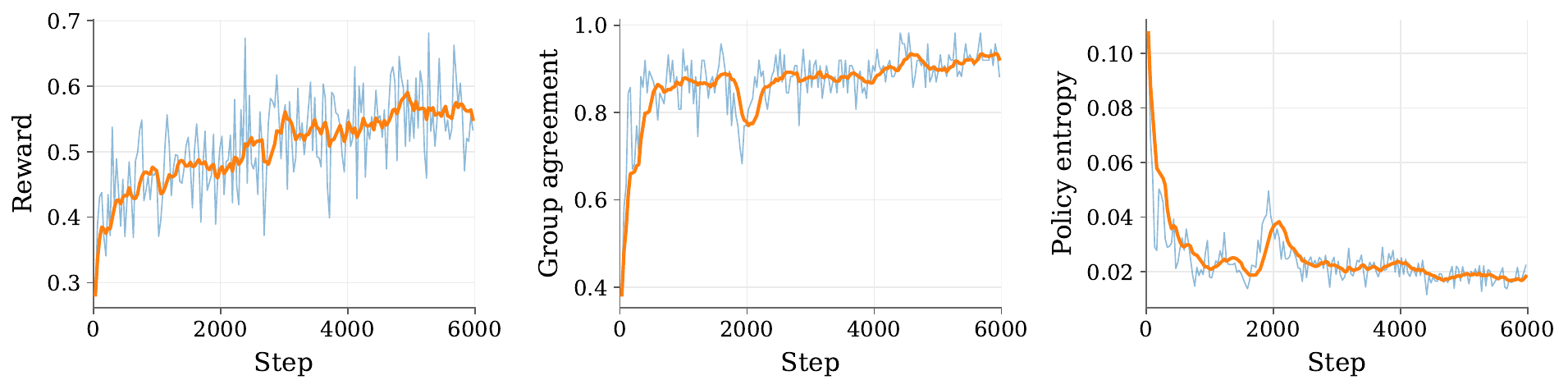}
\caption{\textbf{Training stability of RPSD w/o OT on MovieLens-1M.}
\textbf{(left)} Rule-based reward. \textbf{(middle)} Group agreement. \textbf{(right)} Policy entropy over
completion tokens. Blue curves are the raw logged values and orange curves are moving
averages.}
\end{figure}
 
\begin{figure}[t]
\centering
\includegraphics[width=\textwidth]{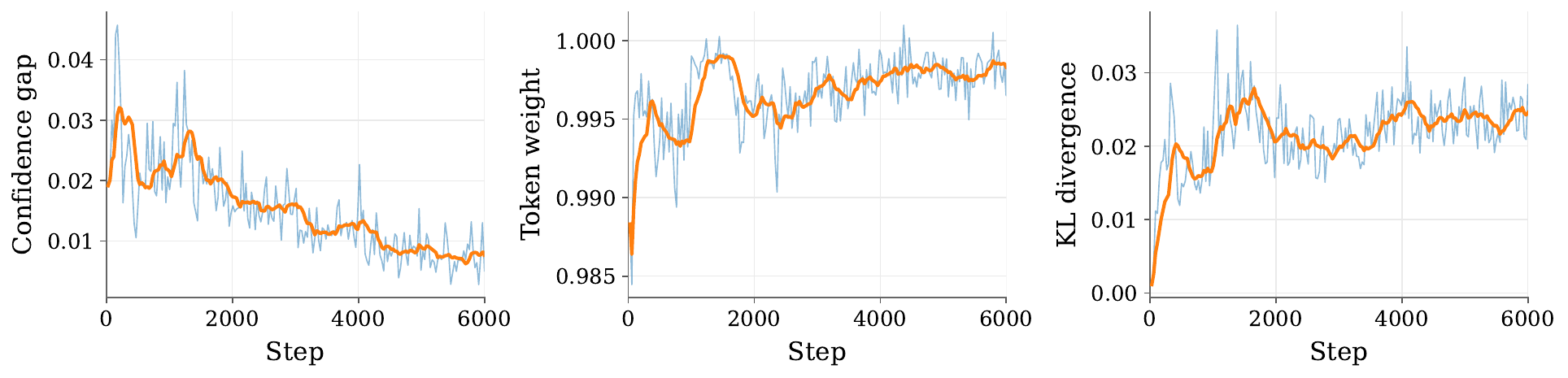}
\caption{\textbf{Training diagnostics of RPSD w/o OT on MicroLens.} The
teacher is privileged with the ground-truth item alone. \textbf{(left)} The
teacher--student confidence gap averaged over completion tokens.
\textbf{(middle)} The advantage-aware token weight. \textbf{(right)}
The KL divergence to the reference policy. Blue curves are the raw
logged values and orange curves are moving averages.}
\end{figure}
 
\begin{figure}[t]
\centering
\includegraphics[width=\textwidth]{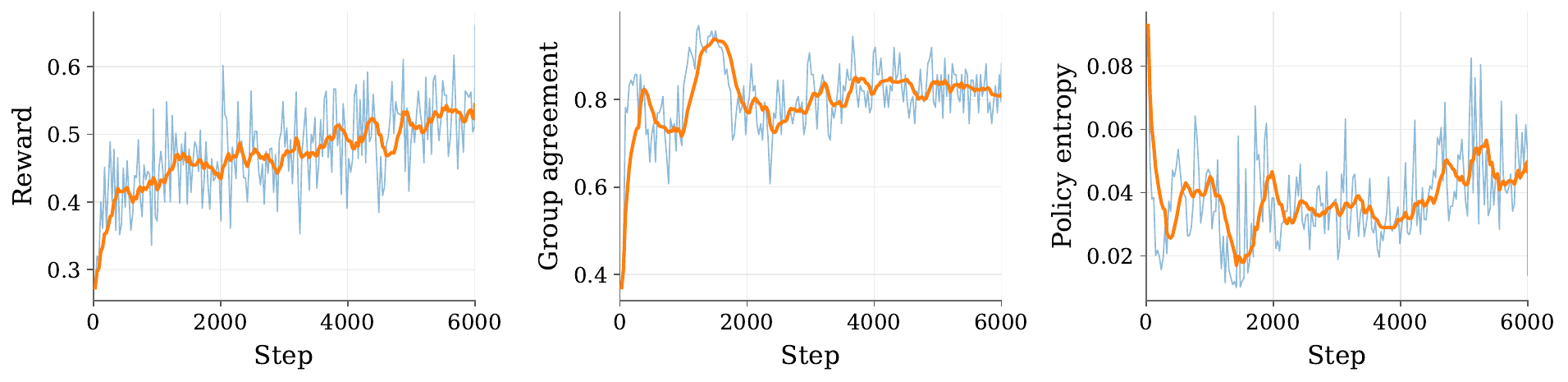}
\caption{\textbf{Training stability of RPSD w/o OT on MicroLens.}
\textbf{(left)} Rule-based reward. \textbf{(middle)} Group agreement. \textbf{(right)} Policy entropy over
completion tokens. Blue curves are the raw logged values and orange curves are moving
averages.}
\end{figure}
 
\begin{figure}[t]
\centering
\includegraphics[width=\textwidth]{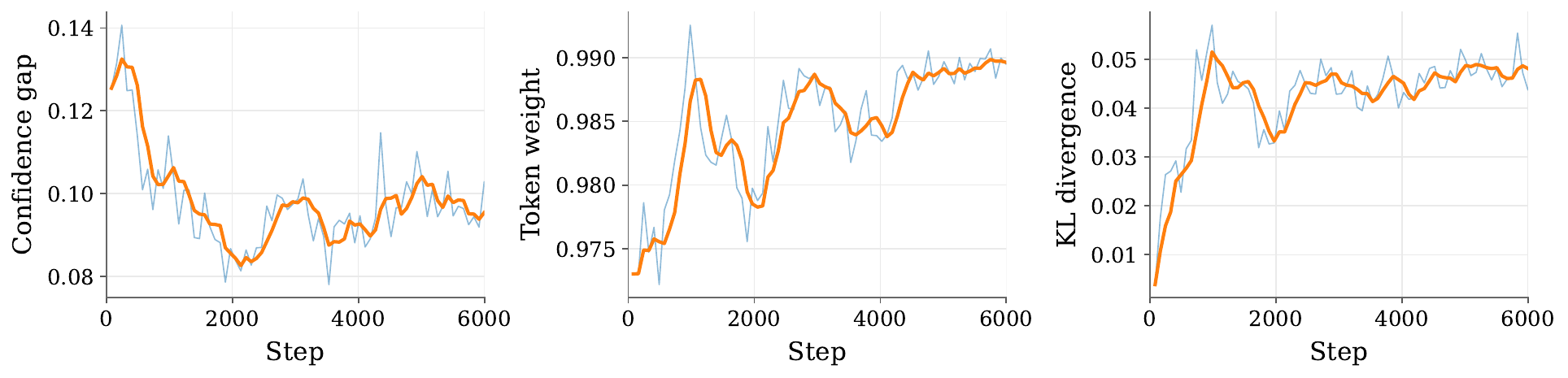}
\caption{\textbf{Training diagnostics of RPSD w/o OT on Netflix.} The
teacher is privileged with the ground-truth item alone. \textbf{(left)} The
teacher--student confidence gap averaged over completion tokens.
\textbf{(middle)} The advantage-aware token weight. \textbf{(right)}
The KL divergence to the reference policy. Blue curves are the raw
logged values and orange curves are moving averages.}
\end{figure}
 
\begin{figure}[t]
\centering
\includegraphics[width=\textwidth]{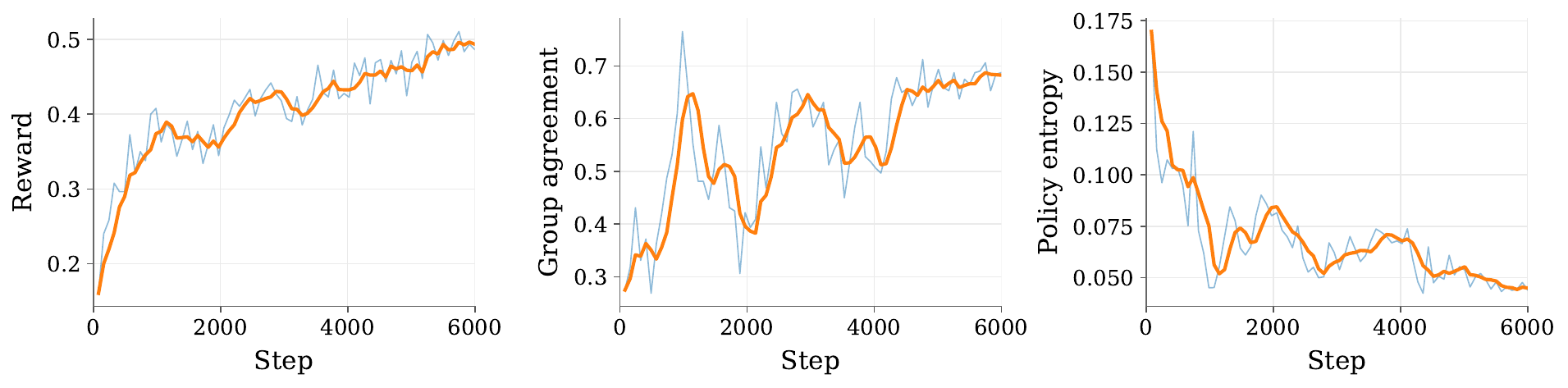}
\caption{\textbf{Training stability of RPSD w/o OT on Netflix.}
\textbf{(left)} Rule-based reward. \textbf{(middle)} Group agreement. \textbf{(right)} Policy entropy over
completion tokens. Blue curves are the raw logged values and orange curves are moving
averages.}
\label{w/o OT last}
\end{figure}

\begin{table}[t]
\centering
\caption{Performance comparison with vanilla OPSD.}
\label{tab:comparison OPSD}
\setlength{\tabcolsep}{25pt}
\renewcommand{\arraystretch}{1.25}
\resizebox{\linewidth}{!}{
\begin{tabular}{l cccc}
\toprule
& \multicolumn{4}{c}{\textbf{MicroLens}} \\
\cmidrule(lr){2-5}
\textbf{Model} & HR@3 & HR@5 & NG@3 & NG@5 \\
\midrule
 \textbf{Qwen3-4B + SFT}      & 0.5187 & 0.6526 & 0.4339 & 0.4982 \\
\textbf{Qwen3-4B + GRPO}      &  0.6621 & 0.7764 & 0.5841 & 0.6330 \\
\textbf{Qwen3-4B + OPSD}      &  0.5669 & 0.6942 & 0.4809 & 0.5404 \\
\textbf{Qwen3-4B + RPSD (Ours)} &  \textbf{0.6836}  & \textbf{0.7809} &\textbf{0.6036} & \textbf{0.6417} \\
\bottomrule
\end{tabular}}
\end{table}

In this section, we examine how the training dynamics change once the retrieved
evidence by OT is removed. In this variant, the teacher is privileged with the
ground-truth item alone and reads no behavioral evidence, so it knows which
candidate is correct but has no account of why it fits the current user history.
Everything else, including the backbone, the prompt, the sampling configuration,
the reward, and the guidance strength, is identical to the full model. The
diagnostics are shown in Figures~\ref{w/o OT first} - \ref{w/o OT last}.

The overall optimization remains stable and close to that of the full model. On
all three datasets, the rule-based reward still rises without collapse, the policy
entropy still decreases, and the KL to the reference stays small and flat. In this
respect the label-only teacher behaves like the full model: it does not destabilize
training or pull the policy away from the reference.

The difference appears in the confidence gap $\delta_{i,t}$, the quantity that
measures how differently the teacher scores the student's tokens. Without behavioral
evidence, the teacher and the student read almost the same input, so on the denser
datasets the teacher rescores the tokens in nearly the same way and $\delta_{i,t}$
stays small, giving the student only a weak token-level correction. On the sparse
Netflix dataset the label-only teacher instead disagrees with the student in an
unstable way, so $\delta_{i,t}$ is larger but noisy. In neither case does removing
the evidence break training and it mainly weakens or destabilizes the teacher--student
correction that reshapes token-level credit, which is the signal the full model
relies on.

How strongly the teacher can still guide the student depends on how much a user's
own history already reveals. On the denser datasets (MovieLens-1M and MicroLens),
where each user has a long history, the confidence gap $\delta_{i,t}$ is small from
the start, because the student can already infer the answer and the label alone provides limited information. On the sparse Netflix dataset, the teacher-student gap opens wider and does not close since the label identifies the
correct item but gives the student no way to reach that conclusion on its own. Therefore, the
teacher and the student keep disagreeing without resolving it. This
matches the ablation study, where removing the retrieved evidence hurts accuracy the
most on the sparsest dataset, and it supports our claim that the benefit of RPSD comes from the behavioral evidence rather than from merely conditioning the teacher on the answer.
\begin{table}[t]
\centering
\caption{Ablation on the Response-level RPSD and Token-level RPSD (Ours).}
\label{response-level}
\setlength{\tabcolsep}{25pt}
\renewcommand{\arraystretch}{1.25}
\resizebox{\linewidth}{!}{
\begin{tabular}{l cccc}
\toprule
& \multicolumn{4}{c}{\textbf{Netflix}} \\
\cmidrule(lr){2-5}
\textbf{Model} & HR@3 & HR@5 & NG@3 & NG@5 \\
\midrule
\textbf{GRPO}      &   0.6690 & 0.8510 & 0.5490 & 0.6198 \\
\textbf{Response-level RPSD}      & 0.6975  & 0.8655 &  0.5775& 0.6439\\
\textbf{Token-level RPSD (Ours)} &   \textbf{0.8020} &  \textbf{0.9245} & \textbf{0.6952} & \textbf{0.7448} \\
\bottomrule
\end{tabular}}
\end{table}

\section{Comparison with Vanilla OPSD}
\label{app:opsd}

Vanilla on-policy self-distillation (OPSD)~\citep{opsd}
uses the same language model as a student and a privileged teacher under
different conditioning contexts. Given an input $x$ and its reference
solution $y^\star$, the student first samples an on-policy response
$\hat{y}\sim p_S(\cdot\mid x)$. The teacher then evaluates the same
student-generated prefixes while additionally conditioning on $y^\star$.
Its training objective minimizes the token-wise divergence between the
full next-token distributions of the teacher and student:
\begin{equation}
\mathcal{L}_{\mathrm{OPSD}}(\theta)
=
\mathbb{E}_{(x,y^\star)\sim\mathcal{D}}
\mathbb{E}_{\hat{y}\sim p_S(\cdot\mid x)}
\left[
\frac{1}{|\hat{y}|}
\sum_{t=1}^{|\hat{y}|}
D\left(
p_T(\cdot\mid x,y^\star,\hat{y}_{<t})
\,\Vert\,
p_S(\cdot\mid x,\hat{y}_{<t})
\right)
\right],
\end{equation}
where $D$ can be instantiated as KL divergence or generalized
Jensen--Shannon divergence. Gradients are propagated only through the
student distribution, while the privileged teacher serves as a detached
distributional target.

RPSD differs from vanilla OPSD in both how the teacher signal is incorporated into optimization and what privileged knowledge the teacher receives. Rather than directly matching the complete student
distribution to the teacher distribution, RPSD retains the
reward-grounded GRPO objective and uses the teacher only to modulate the
magnitude of token-level credit. Specifically, given the group-normalized
response advantage $\widehat{A}_i$, RPSD computes
\begin{equation}
\Delta_{i,t}
=
\log \pi_{\mathrm{sg}(\theta)}
(o_{i,t}\mid x_i^T,o_{i,<t})
-
\log \pi_{\mathrm{sg}(\theta)}
(o_{i,t}\mid x_i^S,o_{i,<t}),
\end{equation}
where the teacher context $x_i^T$ additionally contains the ground-truth
item and retrieved behavioral evidence. After centering and clipping the
confidence gap into $\delta_{i,t}$, the token weight and refined advantage
are defined as
\begin{equation}
w_{i,t}
=
\operatorname{clip}
\left(
\exp\left[
\operatorname{sign}(\widehat{A}_i)\delta_{i,t}
\right],
1-\epsilon_w,
1+\epsilon_w
\right),
\qquad
\widetilde{A}_{i,t}
=
\widehat{A}_i
\left[(1-\lambda)+\lambda w_{i,t}\right].
\end{equation}
Since $w_{i,t}>0$ and $\lambda\in[0,1]$, the scaling factor is strictly
positive and therefore
$\operatorname{sign}(\widetilde{A}_{i,t})
=\operatorname{sign}(\widehat{A}_i)$.
Thus, the verifiable reward continues to determine whether the
policy-gradient term reinforces or suppresses a response, while the
teacher only redistributes its magnitude across tokens. This is fundamentally different from vanilla OPSD, where the teacher distribution
itself constitutes the optimization target. RPSD further differs from vanilla OPSD
in the source of the teacher's privileged information. Vanilla OPSD conditions the
teacher on the ground-truth label alone, which tells the teacher which candidate is
correct but not why it fits the current history. In contrast, RPSD additionally
conditions the teacher on cross-user behavioral evidence retrieved by temporally
aware optimal transport, so the teacher is grounded in how behaviorally similar
users actually continued rather than in the bare answer.

As reported in Table~\ref{tab:comparison OPSD}, vanilla OPSD improves over SFT,
confirming that dense on-policy teacher supervision is a useful signal. However, it
falls below GRPO on the recommendation performance. Directly imitating a label-conditioned teacher distribution pulls every token toward the teacher, including tokens that carry no evidence for the recommendation, so not all updates stay aligned with the verifiable objective. RPSD
avoids this by keeping the reward-grounded objective and letting the teacher only
reweight token credit, so the reward still decides the direction of each update while
the retrieved evidence by OT and the ground-truth item sharpen where that update lands. This is why RPSD surpasses
both OPSD and GRPO rather than sitting between them.

\section{Response-Level versus Token-Level Credit}
\label{app:response_level_credit}

The full RPSD applies the teacher's guidance at the token level. Each token gets its
own weight, so tokens the retrieved evidence supports are reinforced more
and formatting tokens are down-weighted. To test that this token-level granularity is truly the source of the performance
gain, and not just a per-response calibration, we relax RPSD into a response-level
variant that keeps everything else fixed and only changes the granularity at which
the teacher's guidance is applied. The Response-level RPSD shares the same student rollouts,
retrieval-privileged teacher, rule-based rewards, confidence gaps $\delta_{i,t}$, and
hyperparameters as RPSD.

The relaxation averages the per-token weights of each response into a single scalar
and broadcasts it to all tokens,
\begin{equation}
\bar{w}_i = \frac{1}{T_i}\sum_{t=1}^{T_i} w_{i,t},
\qquad
\widetilde{A}^{\mathrm{resp}}_{i,t} = \widehat{A}_i\left[(1-\lambda)+\lambda\bar{w}_i\right],
\quad t=1,\ldots,T_i,
\end{equation}
where $w_{i,t}$ is computed exactly as in the full RPSD. GRPO shares one advantage across all tokens, Response-level RPSD rescales each response using a single teacher-derived scalar, whereas our RPSD maintains a separate weight for each token. Since none of the three methods reverses the sign of $\widehat{A}_i$, the verifiable reward still determines the direction of every update. Therefore, these methods differ only in the granularity with which the teacher redistributes the update magnitude.

Table~\ref{response-level} provides a clear comparison. Response-level RPSD already improves over GRPO, showing that the privileged teacher remains useful even when its guidance is compressed into a single scalar per response. However, it still falls well below the full Token-level RPSD. This suggests that the gain of RPSD does not arise merely from strengthening some responses and weakening others, but largely from distinguishing tokens within each response. This difference follows directly from the weight construction. Because $\delta_{i,t}$ is centered within each response, averaging it largely removes the contrast between above- and below-average tokens, making Response-level RPSD much closer to GRPO's uniform within-response credit. It can still modulate the overall update strength across responses, but cannot distinguish which tokens are more strongly supported by the privileged teacher. In contrast, Token-level RPSD assigns larger updates to tokens whose likelihood changes more under the retrieved evidence while down-weighting less informative tokens, such as generic or formatting tokens. The larger improvement in NDCG further suggests that such fine-grained credit helps not only identify the target item but also rank it more accurately.



\section{Case Study of Student and Teacher Prompts}
\label{app:prompt_showcase}

To make the privileged knowledge concrete, Figure~\ref{fig:prompt_showcase} shows an example of student and teacher prompts. The student prompt $x^S$ contains only the user's
interaction history and the candidate set, and asks the model to name the single next
item. It is the sole input at inference, so the deployed policy never sees any of the
privileged information below.

During training, the teacher additionally reads two blocks. The first is the retrieved
evidence $E_u$, which lists how behaviorally similar users continued after comparable
histories, together with a similarity score for each trajectory. In this example, the
user's recent history mixes classic science-fiction titles such as
\emph{Mad Max}, \emph{The Matrix}, and \emph{Men in Black}. The retrieved
trajectories share several of these items, so their continuations suggest the kind of
item such a user tends to watch next. The second block is the ground-truth next item,
which states the correct answer directly. These privileged views together let the teacher judge the
student's response not only by whether it matches the label, but also in light of what
similar users actually did.

\clearpage
\begin{tcolorbox}[breakable, enhanced, width=\linewidth, colback=gray!4,
    colframe=gray!55, boxrule=0.6pt, arc=2pt, left=5pt, right=5pt, top=4pt,
    bottom=4pt, fonttitle=\bfseries,
    title={Student prompt $x^{S}$ (used at both training and inference)}]
{\footnotesize\ttfamily\baselineskip=11pt
You are a movie recommendation assistant.\\
Here are the user's most recently watched movies sequence:\\
1. [ITEM\_1199] Brazil (1985)\\
2. [ITEM\_750] Dr. Strangelove (1964)\\
3. [ITEM\_924] 2001: A Space Odyssey (1968)\\
4. [ITEM\_1253] Day the Earth Stood Still, The (1951)\\
5. [ITEM\_2660] Thing From Another World, The (1951)\\
6. [ITEM\_3702] Mad Max (1979)\\
7. [ITEM\_2571] Matrix, The (1999)\\
8. [ITEM\_1580] Men in Black (1997)\\[2pt]
Candidate movies:\\
{[ITEM\_926]} All About Eve (1950)\\
{[ITEM\_460]} Getting Even with Dad (1994)\\
{[ITEM\_1142]} Get Over It (1996)\\
{[ITEM\_1016]} Shaggy Dog, The (1959)\\
{[ITEM\_1129]} Escape from New York (1981)\\
{[ITEM\_104]} Happy Gilmore (1996)\\
{[ITEM\_3106]} Come See the Paradise (1990)\\
{[ITEM\_3085]} Living Dead Girl, The (1982)\\
{[ITEM\_575]} Little Rascals, The (1994)\\
{[ITEM\_2688]} General's Daughter, The (1999)\\[2pt]
Now answer the following question.\\
Only output in format [ITEM\_xxxx] Movie Title.\\
No additional caption, summary, comments, or explanation.\par}
\end{tcolorbox}

\begin{tcolorbox}[breakable, enhanced, width=\linewidth, colback=blue!4,
    colframe=blue!45, boxrule=0.6pt, arc=2pt, left=5pt, right=5pt, top=4pt,
    bottom=4pt, fonttitle=\bfseries,
    title={Teacher-only privileged information: evidence block $E_u$ and ground-truth item $v^{u}_{t+1}$ (training only)}]
{\footnotesize\ttfamily\baselineskip=11pt
[Retrieved behavioral evidence]\\
Similar historical trajectories from behaviorally comparable users:
1. [ITEM\_32] $\to$ [ITEM\_198] $\to$ [ITEM\_1580] $\to$ [ITEM\_1374] $\to$ [ITEM\_3702], next [ITEM\_2985] (sim.\ 0.12)

2. [ITEM\_198] $\to$ [ITEM\_1580] $\to$ [ITEM\_1374] $\to$ [ITEM\_3702] $\to$ [ITEM\_2985], next [ITEM\_24] (sim.\ 0.11)

3. [ITEM\_2571] $\to$ [ITEM\_480] $\to$ [ITEM\_1653] $\to$ [ITEM\_1396] $\to$ [ITEM\_1580], next [ITEM\_1527] (sim.\ 0.09)

[Ground-truth next-item evidence]

Correct next item: {[ITEM\_1129]} Escape from New York (1981).

Score the already sampled answer under this evidence. Do not generate a new answer.\par}
\end{tcolorbox}

\begin{tcolorbox}[breakable, enhanced, width=\linewidth, colback=orange!5,
    colframe=orange!55, boxrule=0.6pt, arc=2pt, left=5pt, right=5pt, top=4pt,
    bottom=4pt, fonttitle=\bfseries,
    title={Teacher view $x^{T}=x^{S}\oplus E_u\oplus v^{u}_{t+1}$ (training only)}]
{\footnotesize\ttfamily\baselineskip=11pt
The teacher view is the student prompt above followed by the retrieved
evidence block $E_u$ and the ground-truth next item $v^{u}_{t+1}$.

Conditioned on this privileged information, the teacher re-scores the
same sampled tokens under stop-gradient to produce token-level credit,
and it does not generate a new answer. Neither component is visible to the student at training or at inference.\par}
\end{tcolorbox}

\vspace{3pt}
\captionof{figure}{An example of the student prompt $x^{S}$ and the
teacher's privileged knowledge.}
\label{fig:prompt_showcase}

\end{document}